\documentclass[10pt,twocolumn,letterpaper]{article}

\usepackage{wacv}
\usepackage{times}
\usepackage{epsfig}
\usepackage{graphicx}
\usepackage{amsmath}
\usepackage{amssymb}
\usepackage[accsupp]{axessibility}  

\wacvfinalcopy 

\usepackage[breaklinks=true,bookmarks=false]{hyperref}

\usepackage{booktabs}
\usepackage{soul}
\usepackage{cuted}
\usepackage{caption}
\usepackage{subcaption}
\usepackage[dvipsnames]{xcolor}

\DeclareMathOperator*{\argmin}{arg\,min} 

\newcommand{\fig}[5]{
	\begin{figure}[#1]
		\begin{center}
			\includegraphics*[#2]{#3}
			\caption{#5}
			\label{#4}
		\end{center}
	\end{figure}
}

\newcommand{\figfw}[5]{
	\begin{figure*}[#1]
		\begin{center}
			\includegraphics*[#2]{#3}
			\caption{#5}
			\label{#4}
		\end{center}
	\end{figure*}
}
\begin{document}

\title{PRECODE - A Generic Model Extension to Prevent Deep Gradient Leakage}

\author{Daniel Scheliga$^1$ \and Patrick M\"{a}der$^{1,2}$ \vspace{2pt} \and Marco Seeland$^1$\\

\\$^1$Technische Universit\"{a}t Ilmenau, $^2$Friedrich Schiller Universit\"{a}t Jena, Germany\\
{\tt\small \{daniel.scheliga,patrick.maeder,marco.seeland\}@tu-ilmenau.de}
}

\maketitle

\begin{abstract}
Collaborative training of neural networks leverages distributed data by exchanging gradient information between different clients.
Although training data entirely resides with the clients, recent work shows that training data can be reconstructed from such exchanged gradient information.
To enhance privacy, gradient perturbation techniques have been proposed.
However, they come at the cost of reduced model performance, increased convergence time, or increased data demand.
In this paper, we introduce PRECODE, a PRivacy EnhanCing mODulE that can be used as generic extension for arbitrary model architectures.
We propose a simple yet effective realization of PRECODE using variational modeling.
The stochastic sampling induced by variational modeling effectively prevents privacy leakage from gradients and in turn preserves privacy of data owners.
We evaluate PRECODE using state of the art gradient inversion attacks on two different model architectures trained on three datasets.
In contrast to commonly used defense mechanisms, we find that our proposed modification consistently reduces the attack success rate to $0\%$ while having almost no negative impact on model training and final performance.
As a result, PRECODE reveals a promising path towards privacy enhancing model extensions.
\end{abstract}

\section{Introduction}
\label{sec:intro}

In application domains like medical imaging, data sharing is typically prohibited due to privacy, legal, and technical regulations~\cite{li2019privacy}.
Avoiding the need for shared data, collaborative training such as federated learning allows multiple clients to collaboratively compute a common model~\cite{mcmahan2017communication}.
This collaborative use of distributed data can result in a severe boost of model performance, while mitigating a lot of systemic privacy risks and costs inherent in model training on centrally aggregated data~\cite{kairouz2019advances}.

\begin{figure}
     \centering
     \begin{subfigure}{.24\linewidth}
         \centering
         \includegraphics[width=.9\linewidth]{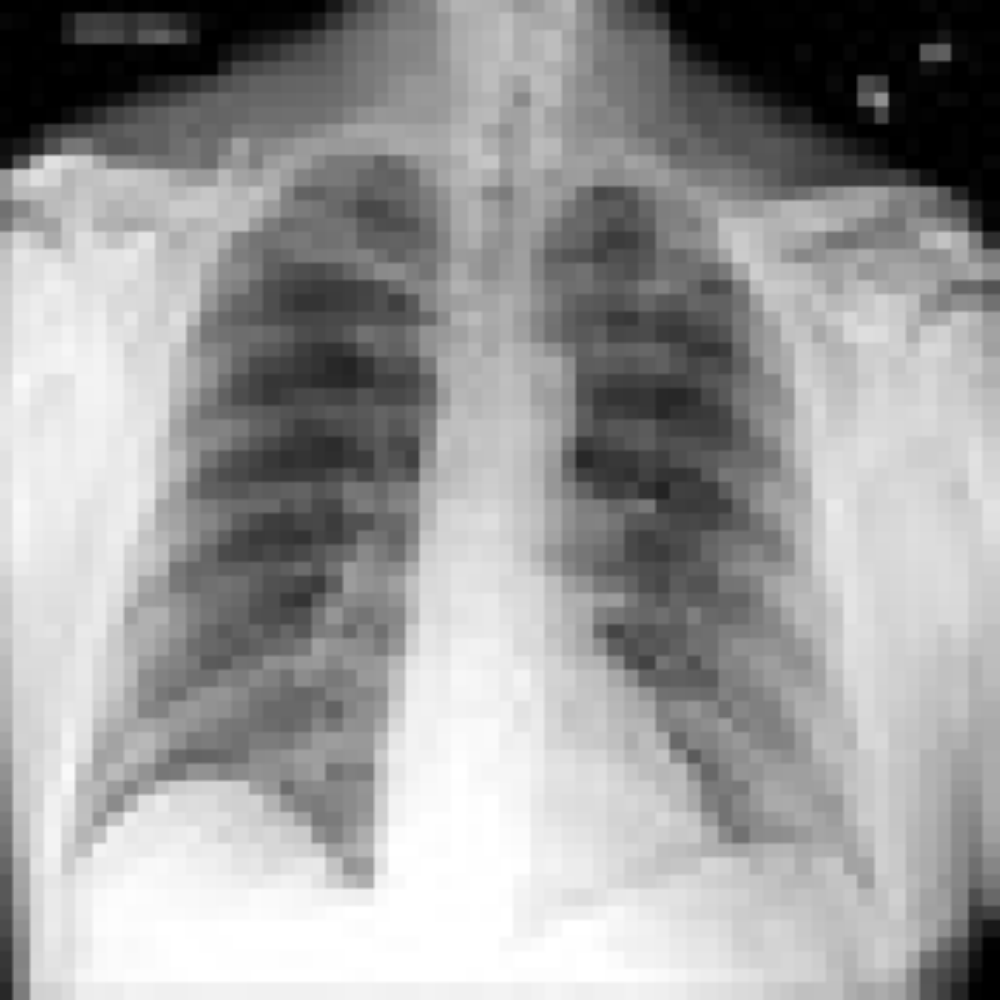}  
         \vspace{-4pt}
         \caption{}
     \end{subfigure}
     \begin{subfigure}{.24\linewidth}
         \centering
         \includegraphics[width=.9\linewidth]{images/MedMNIST_orig_lung.pdf}
         \vspace{-4pt}
         \caption{}
     \end{subfigure}
     \begin{subfigure}{.24\linewidth}
         \centering
         \includegraphics[width=.9\linewidth]{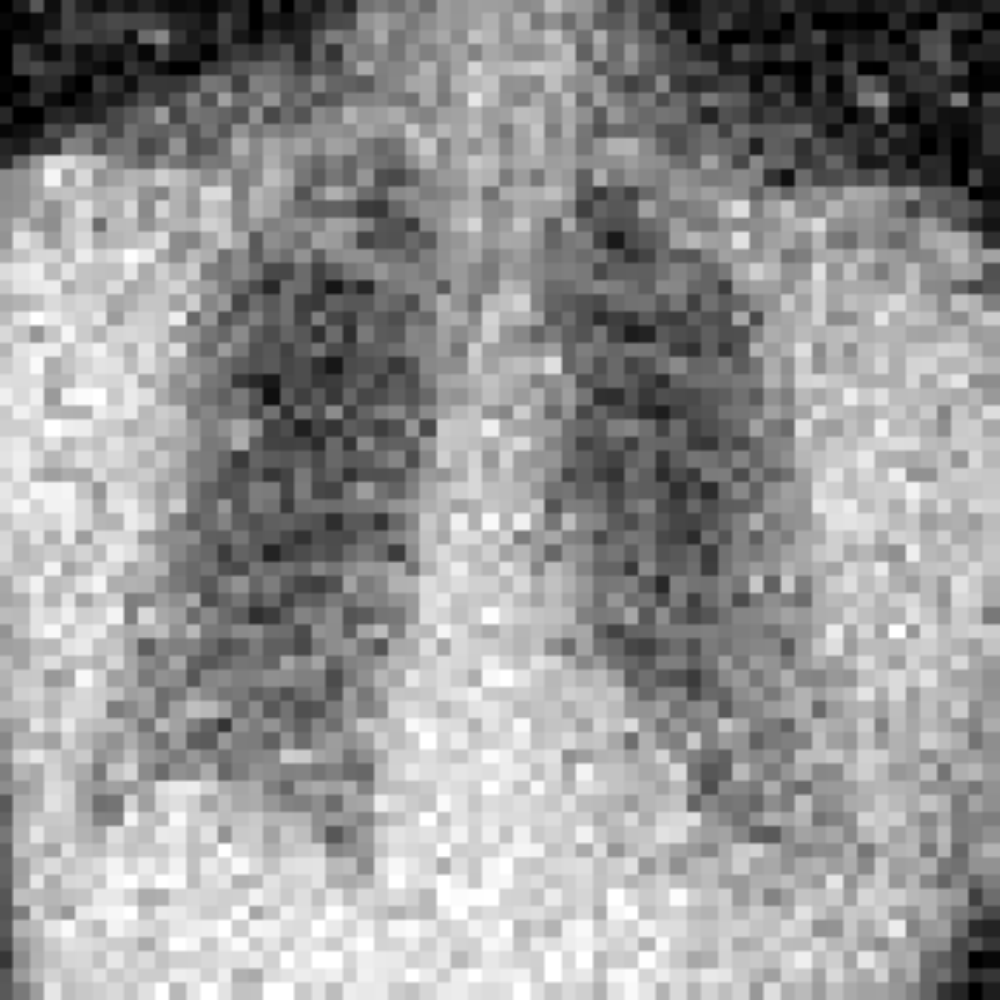}  
         \vspace{-4pt}   
         \caption{}
     \end{subfigure}
     \begin{subfigure}{.24\linewidth}
         \centering
         \includegraphics[width=.9\linewidth]{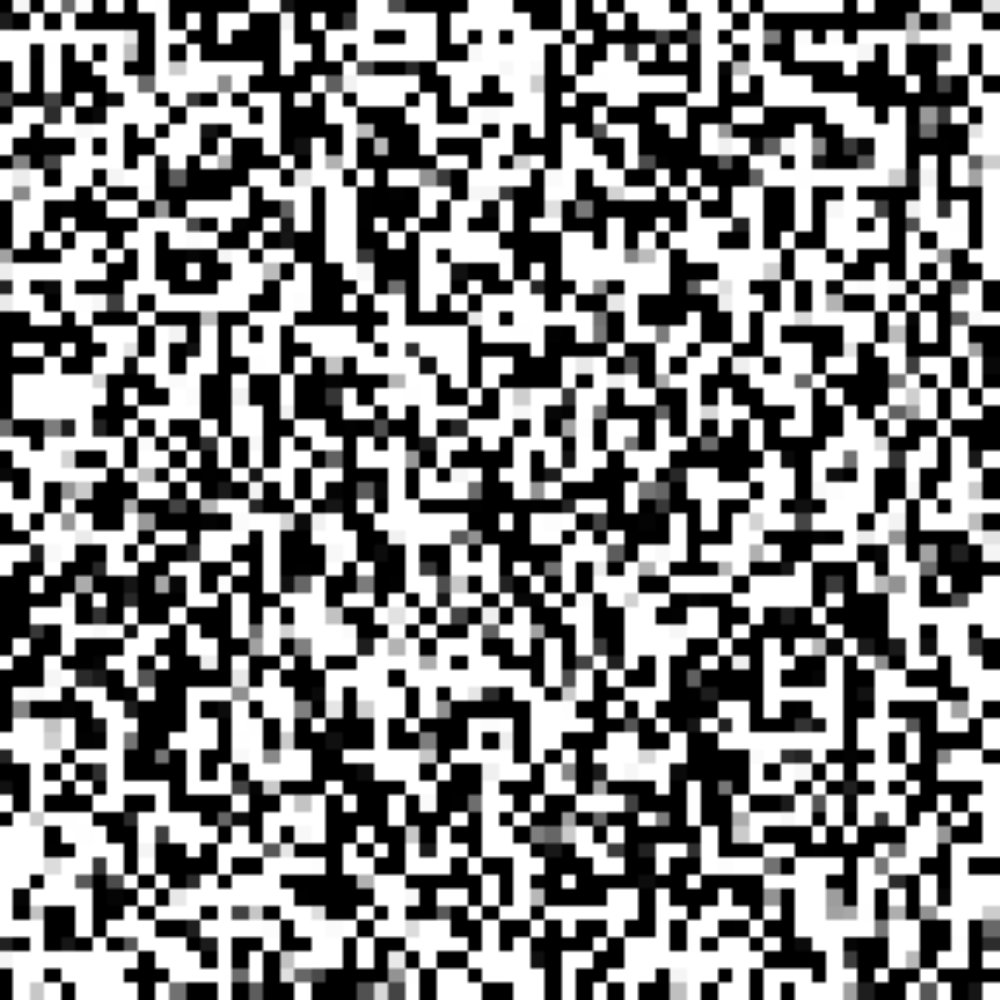}  
         \vspace{-4pt}
         \caption{}
     \end{subfigure}
     \vspace{-6pt}
     \setlength{\belowcaptionskip}{-8pt}
     \caption{\textbf{Privacy leakage by reconstructing training data from gradients.} (a) Original image. (b) Reconstructed image. (c) Image reconstructed from perturbed gradient information. (d) Prevented reconstruction using our proposed privacy-enhanced model architecture.}
     \label{fig:teaser}
\end{figure}

Although training data entirely resides with the respective clients, recent studies demonstrated the possibility for compromising clients' privacy by reconstructing sensitive information from exchanged model parameters.
The most advanced techniques for such privacy leakage are gradient inversion attacks~\cite{geiping2020inverting, wei2020framework, zhao2020idlg, zhu2019deep} that allow to reconstruct training data from exchanged model weights or gradients.
Exemplary reconstructions are shown in Fig.~\ref{fig:teaser}.
Geiping \etal~\cite{geiping2020inverting} show that the input to any fully connected layer in a neural network can be analytically reconstructed given the model architecture and associated gradients.
As basically all state of the art deep learning architectures make use of fully connected layers, the privacy of individual data owners is clearly exposed to such attacks.

Since such privacy leaks are based on exchanged gradients,~\cite{wei2020framework, zhu2019deep} propose to perturb gradients before sharing.
As a side effect an inherent trade-off between privacy and model performance can be observed~\cite{jayaraman2019evaluating, wei2020framework, zhu2019deep}.

To avoid this trade-off, we propose to enhance the architecture of arbitrary models by inserting PRECODE, a \textit{PRivacy EnhanCing mODulE}. 
The idea of PRECODE is to use variational modeling to disguise the original latent feature space vulnerable to privacy leakage by gradient inversion attacks.
We perform a systematic evaluation to investigate the privacy enhancing properties of our proposed model extension.
In contrast to gradient perturbation based defense mechanisms, we show that PRECODE has minimal impact on the training process and model performance.

\begin{figure}[h]
    \centering
    \includegraphics[width=\linewidth]{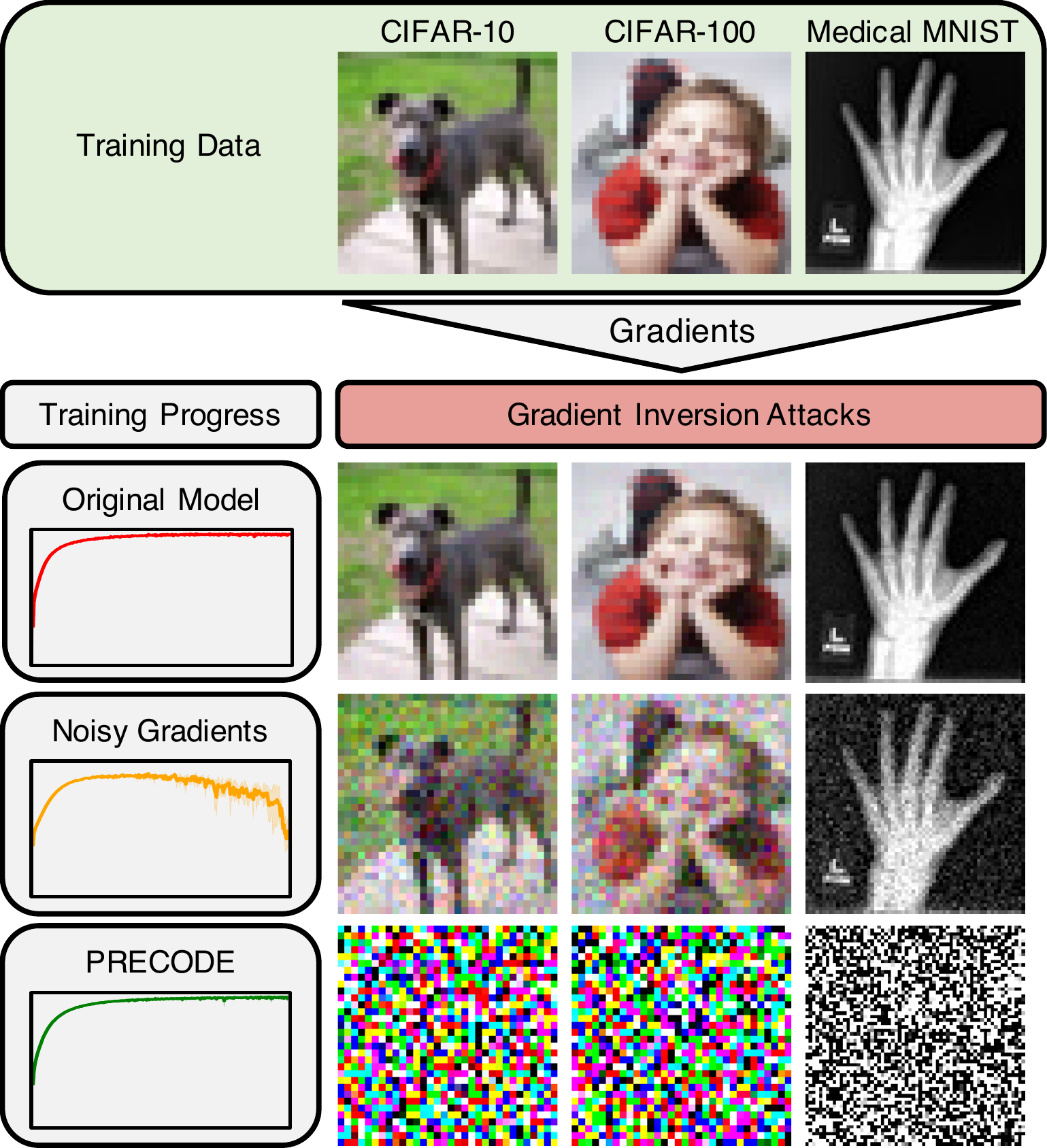}
    \caption{
\textbf{Summary of the content of this paper.}
Neural networks are trained on CIFAR-10, CIFAR-100, and Medical MNIST.
During training, gradients are used for inversion attacks aiming to reconstruct original training data.
Without defense (Original Model), training data can be entirely reconstructed.
Noisy gradients prevent reconstruction to some extent but reduce model performance.
Our proposed PRECODE extension entirely prevents reconstruction without impacting the training process.
}
    \label{fig:attack_samples}
\end{figure}

Fig.~\ref{fig:attack_samples} displays a visual summary of the content of this paper.
Our contributions can be summarized as follows:
\begin{itemize}
    \setlength{\topsep}{-2pt}
    \setlength{\itemsep}{0pt}
    \item To the best of our knowledge, we are first to propose specific model extensions to prevent privacy leakage.
    \item We propose a generic privacy enhancing module for defending gradient inversion attacks.
    \item We perform a systematic empirical evaluation using state of the art gradient inversion attacks on two model architectures trained on three image classification datasets and compare our method with two state of the art defense methods.
\end{itemize}


\section{Related Work}
\label{sec:sota}

\subsection{Gradient Inversion Attacks} 
\label{sec:sota_rec_att}
Gradient inversion attacks as studied in \cite{geiping2020inverting, wei2020framework, zhao2020idlg,zhu2019deep} aim to reconstruct input training data $x$ from their corresponding gradients $\nabla \mathcal{L}_\theta(\hat{y}, y)$, where $\theta$ are the parameters of the attacked model $F$, $\hat{y}=F(x)$ is the models prediction given the input $x$, $y$ the expected output for $x$ and $\mathcal{L}$ the loss function used to optimize $\theta$.

We follow the threat model described by \cite{geiping2020inverting}, \ie, an \textit{honest-but-curious} server.
In detail, an attacker has insight into the training process, \ie, knowledge of $\mathcal{L}$, $F$, $\theta$ and $\nabla \mathcal{L}_\theta(\hat{y}, y)$, but does not actively interfere with it (\textit{honesty}). 
Instead, the attacker aims to reconstruct training data $(x,y)$ of another client participating in the training process (\textit{curiosity}).
This scenario is particularly relevant for collaborative training processes such as \textit{federated learning}~\cite{kairouz2019advances, mcmahan2017communication, shokri2015privacy}, where a shared model is collaboratively trained by participating clients.
Data is deliberately not shared, but remains with each respective client.
A central aggregator distributes a global model to clients, which in turn respond with model updates, \ie, gradients, after local training.
The clients' gradients are then aggregated in order to update the global model.
This process is iteratively repeated until convergence or another termination criterion is reached.
Apart from star-based connectivity pattern with a central aggregator, other connectivity and aggregation patterns such as peer-to-peer or cluster-based collaborative training have been proposed within the federated learning paradigm~\cite{duan2020self, lee2020tornadoaggregate, roy2019braintorrent}
Since gradient information must also be exchanged in these approaches, they suffer from similar vulnerabilities.

Using knowledge about $\mathcal{L}$, $F$, $\theta$ and $\nabla \mathcal{L}_\theta(\hat{y}, y)$, gradient inversion attacks can generally be formulated as an optimization problem.
As described for \textit{Deep Leakage from Gradients} (DLG)~\cite{zhu2019deep}, the basic idea is to feed a random dummy image $x'$ into a model $F$ to obtain a dummy gradient $\nabla \mathcal{L}_\theta(F(x'),y')$ by comparison to a randomly initialized dummy label $y'$.
To reconstruct both the original input data $x$ and the associated output $y$, the Euclidean distance
\begin{equation}
    \label{eq:DLG}
    \argmin_{ \left( x',y' \right) } ||\nabla \mathcal{L}_\theta(F(x), y) - \nabla \mathcal{L}_\theta(F(x'),y')||^2 
\end{equation}
between the attacked gradient and the dummy gradient is iteratively minimized using a L-BFGS optimizer~\cite{liu1989limited}.
In \textit{improved DLG} (iDLG)~\cite{zhao2020idlg}, Zhao \etal found that ground-truth labels $y$ can be analytically reconstructed from gradient information. 
Without the need to optimize for $y'$ in Eq.~\ref{eq:DLG}, optimization becomes faster and more stable.
Wei \etal propose the \textit{Client Privacy Leakage} (CPL) attack by extending the optimization objective of iDLG by an additive label-matching regularization term
\begin{equation}
    \alpha ||F(x') - y||^2
    \label{eq:CPL}
\end{equation}
and found it to further stabilize the optimization~\cite{wei2020framework}.

Using \textit{inverting gradient attacks} (IGA), Geiping \etal~\cite{geiping2020inverting} further improve the reconstruction process by disentangling gradient magnitude and direction.
They propose to minimize cosine distance instead of Euclidean distance between attacked and dummy gradients in Eq.~\ref{eq:DLG} while using the total variation~\cite{rudin1992nonlinear} of the dummy image $x'$ as regularization to constrain the reconstruction to relevant image parts.
In addition, they found the Adam optimizer~\cite{kingma2015adam} to generally yield more sophisticated reconstruction results compared to the L-BFGS optimizer, especially for deeper models and trained models with overall smaller gradient magnitude.


\subsection{Defense Mechanisms}
\label{sec:sota_def}
The authors of CPL~\cite{wei2020framework} present a comprehensive analysis of privacy leakage and deduced relevant parameters and potential mitigation strategies.
They found that batch size, image resolution, choice of activation functions, and the number of local iterations before gradient exchange can impact privacy leakage.
Supporting findings are reported in \cite{geiping2020inverting, pan2020theory, zhao2020idlg, zhu2021r, zhu2019deep}.
Also the training progress was found to impact privacy leakage from gradients, as models trained for more epochs typically yield smaller gradients compared to earlier in the training process~\cite{geiping2020inverting, wang2020sapag}.
Whereas such parameters and conditions are certainly relevant and could be carefully selected to prevent attacks, they give no guarantee that reconstruction of sensitive data by gradient inversion attacks is not possible~\cite{geiping2020inverting}.
In fact, Geiping \etal showcased successful attacks on deep neural networks (ResNet-152) trained for multiple federated averaging steps and even for batches of $100$ images~\cite{geiping2020inverting}.
In addition, parameter selection is often controlled by other factors, such as model and/or hardware limitations.
In conclusion, data privacy has to be actively controlled by purposeful defense mechanisms.

Cryptographic approaches like homomorphic encryption~\cite{aono2017privacy} and secure aggregation schemes~\cite{bonawitz2017practical} potentially allow for the best privacy preservation.
However, they come at extreme computational costs and are not applicable in every collaborative learning scenario~\cite{zhu2019deep}.

Instead of costly cryptographic approaches, most state of the art methods defend against privacy leakage by immediate perturbation of exchanged gradients.
Therefore, three approaches are typically utilized: \textit{gradient quantization}, \textit{gradient compression}, and \textit{noisy gradients}.

\textit{Gradient quantization} is primarily used to reduce communication costs and memory consumption during collaborative training~~\cite{konevcny2016federated, sattler2019robust}.
For quantization, fixed ranges of numerical values are compressed to sets of values, reducing the entropy of the quantized data.
As a side effect, the success of inversion attacks on quantized gradients is also reduced.
Yet the authors of~\cite{zhu2019deep} found that half-precision quantization~\cite{tagliavini2018transprecision} fails to protect training data. 
Only low-bit Int-8 quantization was able to defend against inversion attacks, but also caused a $22.6\%$ drop in model accuracy~\cite{zhu2019deep}.

\textit{Gradient compression/sparsification} by pruning was also introduced to reduce the communication costs during collaborative training~\cite{lin2017deep}.
Here elements of the gradient that carry the least information, \ie, have the smallest magnitude, are pruned to zero.
Like quantization, this reduces the entropy of the gradient and decreases reconstruction success by inversion attacks.
Zhu \etal found a pruning ratio of at least $20\%$ sufficient to suppress the reconstruction success of the input data~\cite{zhu2019deep}.

The idea of using noise to limit the information disclosure about individuals was originally introduced in the field of differential privacy~\cite{abadi2016deep, dwork2014algorithmic, kairouz2019advances}.
State of the art methods use \textit{noisy gradients} to guarantee a provable degree of privacy to clients participating in collaborative training~\cite{bonawitz2017practical, li2019privacy, mcmahan2018learning}.
In practice, Gaussian or Laplace noise is added to the gradients prior to their exchange.
Whereas added noise can suppress gradient inversion attacks, it also negatively impacts both the training process and the final model performance.
Wei \etal demonstrate that privacy leakage from gradients can be prevented at the cost of a severe drop in model accuracy of up to $9.8\%$~\cite{wei2020framework}.
Such performance drop can be mitigated by an increase of training data~\cite{dwork2014algorithmic} which is typically not feasible in practical scenarios.


\section{PRECODE - Privacy Enhancing Module}
\label{sec:method}
As discussed above, commonly used defense mechanisms rely on gradient perturbations.
Although these mechanisms are shown to reduce privacy leakage by gradient inversion attacks, they induce an inherent trade-off between privacy and model performance.
Apart from increased model depth, batch size, and image resolution, to the best of our knowledge, no deliberate architectural modifications have been proposed to reduce privacy leakage.
Hence, we propose to extend neural network architectures with a PRivacy EnhanCing mODulE (PRECODE).
Next, we summarize the key requirements for such module.


\subsection{Requirements for PRECODE}
\label{sec:requirements}
\begin{description}
    \setlength{\itemsep}{0pt}
    \item[\textbf{Privacy enhancement}] The PRECODE extension needs to sufficiently disguise training data information within gradients in order to prevent privacy leakage by gradient inversion attacks.
    \item[\textbf{Generic model integrability}] Targeting a generalized solution, the PRECODE extension should not be limited to certain model architectures but be applicable to arbitrary model architectures. For ease of integration and joint optimization, we also aim for direct integration of the extension into an existing model architecture without requiring further model modifications.
    \item[\textbf{Model performance}] For such an extension to be practically relevant, it should not harm the final model performance or the training process, \eg, by causing increased convergence times.
\end{description}


\subsection{Realization using Variational Modeling}
\label{sec:PRECODE}
\begin{figure}[!t]
	\begin{center}
		\includegraphics*[width=.82\linewidth]{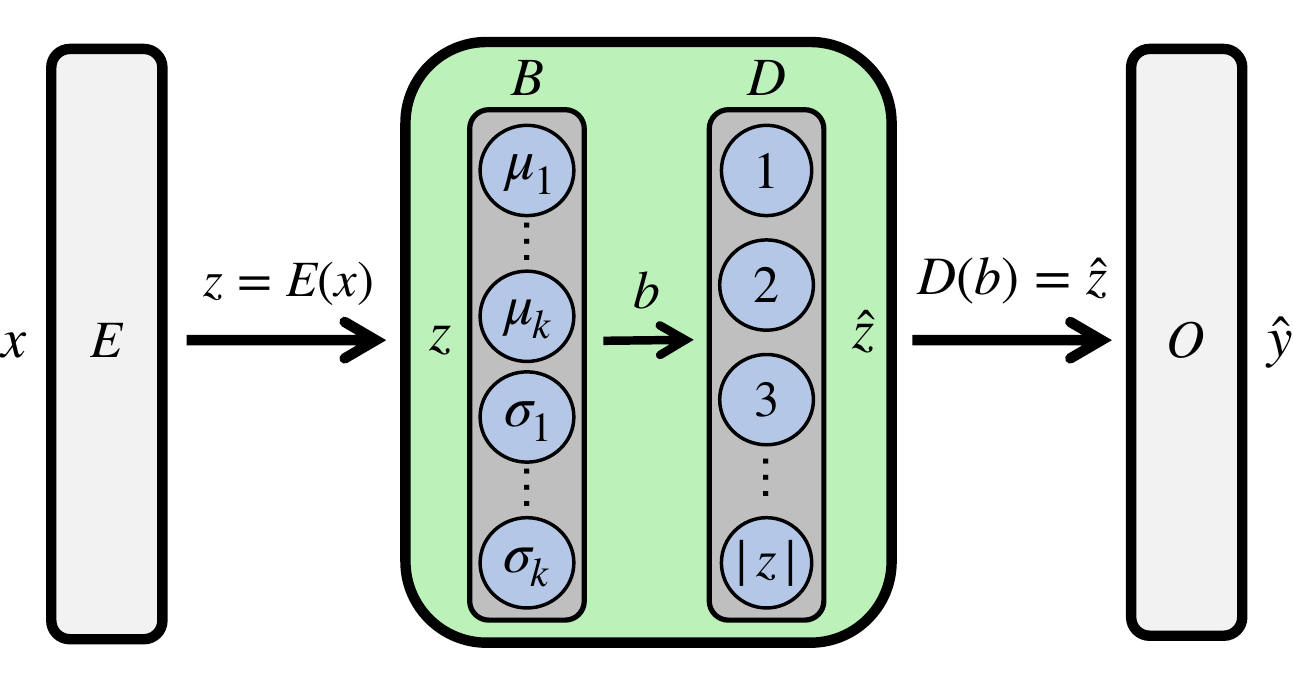}
		\setlength{\belowcaptionskip}{-15pt}
		\caption{Realization of the PRECODE extension as variational bottleneck.}
		\label{fig:vb}
	\end{center}
\end{figure}

Analyzing the criteria described in section~\ref{sec:requirements} and inspired by the work of~\cite{alemi2016deep, kingma2013auto}, we propose the use of a variational bottleneck (VB) as realization of the PRECODE extension.
Aiming to learn a joint distribution between input data and its latent representation, VBs originate from the field of generative modeling~\cite{kingma2013auto}.
They consist of a probabilistic encoder $B=q(b|z)$ and decoder $D=p(z|b)$. 
Fig.~\ref{fig:vb} visualizes the VB and corresponding input and output variables.
We use the VB to approximate the distribution of the latent space and obtain a new representation $\hat{z}$ from an approximated prior distribution $p(b)$ by stochastic sampling.

Hence, the privacy enhancing properties of the VB originate from the fact that the returned latent representation $\hat{z}$ is a \textit{stochastic function} of the input representation $z$.
As $\hat{y}$ and $\mathcal{L}(\hat{y}, y)$ are computed based on this stochastic representation, the gradient $\nabla \mathcal{L}(\hat{y}, y)$ does not contain direct information on the input sample $x$.
Furthermore, iterative optimization-based attacks are countered by design, since $z'=E(x')$ is sampled randomly in each optimization step.
Small changes in $x'$ cause increased entropy of $z'$, therefore making it difficult for the optimizer to find dummy images $x'$ that minimize the reconstruction loss.

We realize $B$ as a fully connected layer with $2k$ neurons, where $k$ defines the size of the bottleneck representation $b$.
Input to the encoder are the hidden representations $z=E(x)$ computed by forward propagation of an input sample $x$ through all layers $E$ of the base neural network prior to the output layer.
$B$ then encodes the representation $z$ into a latent distribution $B(z) = \mathcal{N}(\mu_B, \sigma_B)$ approximated as multivariate Gaussian distribution. 
The bottleneck features $b \sim q(b|z)$ are fed into the stochastic decoder $D$, also realized as fully connected layer.
The number of neurons in $D$ is equal to the number of features in $z$.
$D$ generates a new representation $\hat{z}=p(z|b)p(b)$ which is then used to calculate the model prediction $\hat{y} = O(\hat{z})$.
$O$ corresponds to the layer(s) of the neural network that follow the VB.

For the VB to learn a continuous and complete latent feature space distribution, the loss function $\mathcal{L}_F$ of the base model $F$ is extended by the Kullback-Leibler divergence ($D_{KL}$) between $\mathcal{N}(\mu_B, \sigma_B)$ and a standard normal distribution where $\beta$ controls the weight of the VB loss on the overall loss function:
\begin{equation}
    \label{eq:vb_loss}
    \mathcal{L}(\hat{y}, y) = \mathcal{L}_F(\hat{y},y) + \beta \cdot D_{KL}(\mathcal{N}(\mu_B, \sigma_B), \mathcal{N}(0, 1)).
\end{equation}
In order to backpropagate a gradient through the bottleneck layer, the reparameterization technique described in \cite{kingma2013auto} is used.
We discuss the impact of hyperparameter selection and placement of the VB in the \textit{supplementary material}.
We provide a PyTorch implementation of PRECODE\footnote{\href{https://github.com/SECSY-Group/PRECODE}{https://github.com/SECSY-Group/PRECODE}}. 


\section{Experimental Setup}
\label{sec:exp_setup}
This section describes the experimental settings of the systematic empirical studies conducted to evaluate the behaviour of the proposed PRECODE realization on different model architectures, datasets, and attacks, with respect to model performance and privacy of training data.

We evaluate privacy leakage by reconstructing training images using the state of the art gradient inversion attacks IGA~\cite{geiping2020inverting} and CPL~\cite{wei2020framework}.
For comparative evaluation of PRECODE, we also defend inversion attacks using gradient perturbation techniques, \ie, noisy gradients~\cite{abadi2016deep} and gradient compression~\cite{lin2017deep}.
Gradient quantization is not considered due to the comparatively large negative impact on model performance~\cite{zhu2019deep}.
Model performance is evaluated by training both fully connected as well as convolutional neural networks for image classification on three different datasets, \ie, CIFAR-10, CIFAR-100, and Medical MNIST.

The CIFAR-10 dataset~\cite{krizhevsky2009learning} is composed of $60'000$ colored images with a size of $32 \times 32$ pixels.
Each image displays one out of ten classes and the amount of images per class is equally distributed.
CIFAR-100~\cite{krizhevsky2009learning} represents a harder variant, using same amount of images and same image size but displaying $100$ different classes.
Both datasets are separated into training and test splits according to the corresponding benchmark protocols.
The Medical MNIST dataset \cite{apolanco2017med} comprises six classes of $64 \times 64$ sized grayscale images.
Whereas one class contains $8'954$ images, each other class is represented by $10'000$ images.
The dataset is randomly split into $80\%$ training and $20\%$ test split.

We consider three baseline model architectures with increasing complexity: a shallow MLP (SMLP), deep MLP (DMLP), and LeNet.
MLPs are especially considered as Geiping \etal theoretically proof that the input to fully connected layers can generally be reconstructed from their gradient~\cite{geiping2020inverting}.
The MLPs are composed of fully connected layers with $1024$ neurons and ReLU activation.
Whereas the SMLP uses two hidden layers, the DMLP uses four hidden layers.
The output layer is a fully connected layer with softmax activation, and the number of neurons is equal to the number of classes within a dataset.
In addition to MLPs, we use the LeNet implementation from~\cite{zhao2020idlg} as prototypical architecture for convolutional neural networks.
To investigate the privacy enhancing properties of our proposed method, we extend these baseline architectures by placing PRECODE with $k=256$ and $\beta=0.001$ between the last feature extracting layer and the final classification layer.

To evaluate the model performance and convergence time we train all randomly initialized models for $300$ epochs while minimizing the cross-entropy loss using Adam~\cite{kingma2015adam} optimizer with learning rate $0.001$, momentum parameters $\beta_1=0.9$ and $\beta_2=0.999$, and a batch size of $64$.
For the PRECODE extended models, the loss function is adjusted according to Eq.~\ref{eq:vb_loss}.
Training and test accuracy are measured after each epoch as average across all train/test samples, respectively.
For experiments with gradient perturbation defense mechanisms, the gradients of the cross-entropy loss are perturbed in each optimization step.
All reported metrics are averaged across three runs with different seeds.

For evaluating privacy leakage, we randomly sample a \textit{victim dataset} composed of $128$ images from training data.
We ensured that each class is represented in the victim dataset.
We follow the attack setup as in~\cite{wei2020framework, zhao2020idlg, zhu2019deep} and simulate a single training step to generate the victim gradient using corresponding victim data. 
For reconstruction we use the publicly available PyTorch implementation of IGA\footnote{\href{https://github.com/JonasGeiping/invertinggradients}{https://github.com/JonasGeiping/invertinggradients}}.
We use the default configuration for the attack, \ie, initialization of dummy images from Gaussian distribution, cosine distance based loss function with $\alpha=10^{-6}$ as total variation regularization weight, Adam optimizer with learning rate $0.01$, and step-wise learning rate decay~\cite{geiping2020inverting}.
We found that high quality reconstructions can be generated within $7'000$ iterations.
We stopped attacks if there is no decrease in reconstruction loss for $1'200$ iterations.

Following the findings in \cite{geiping2020inverting, wei2020framework, zhao2020idlg}, we assume label information for each reconstruction to be known as it can be analytically reconstructed from gradients of cross-entropy loss functions w.r.t. weights of fully connected layers with softmax activation (cf. section~\ref{sec:sota_rec_att}).
Please note that the adjusted loss function of PRECODE might impact such analytical label reconstruction.
However, we assume label information also to be known when attacking PRECODE enhanced models.
Otherwise our proposed method might have systematic advantages if attacks would need to jointly optimize for images and their labels.

Similar to~\cite{geiping2020inverting, wei2020framework}, we compute \textit{mean squared error} (MSE), \textit{peak signal-to-noise ratio} (PSNR) and \textit{structural similarity} (SSIM)~\cite{wang2004image} between original and reconstructed images to measure the reconstruction quality.
Better reconstruction, \ie, higher similarity to original images, is indicated by lower MSE as well as higher PSNR and SSIM values.
In addition, we compute the \textit{attack success rate} (ASR) as fraction of the successfully reconstructed images of the victim dataset~\cite{wagner2018technical}.
We consider images as being successfully reconstructed if SSIM $\geq 0.6$.
Using three different seeds, we repeat all inversion attacks on constant sets of $128$ victim images per dataset and report the average MSE, PSNR, SSIM, and ASR.

\section{Results}
\label{sec:results}

\subsection{Network Complexity}
\label{sec:results_net_complexity}
\begin{figure}[!t]
	\begin{center}
		\includegraphics*[width=0.96\linewidth]{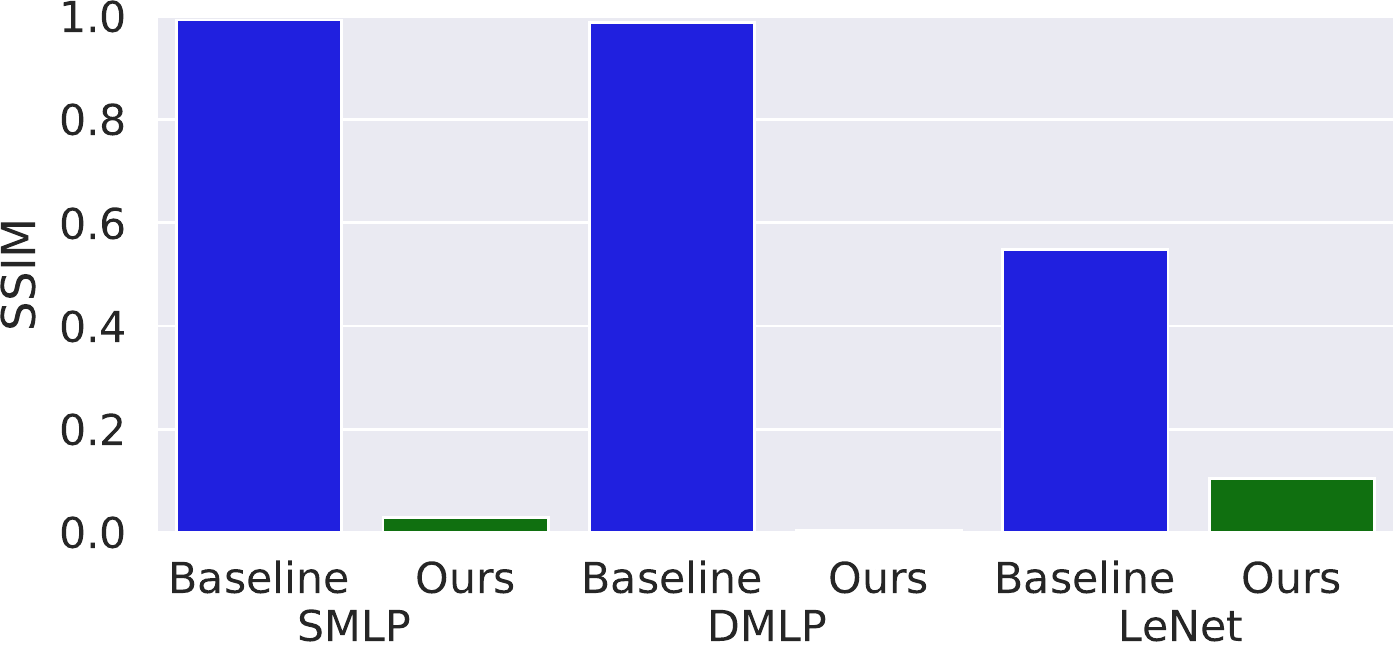}
		\setlength{\belowcaptionskip}{-15pt}
		\caption{SSIM of reconstructed images by IGA on the SMLP, DMLP, and LeNet models for the CIFAR-10 dataset.
                ``Baseline'' denotes original models whereas ``Ours'' denotes extension with PRECODE.}
		\label{fig:ssim_e1}
	\end{center}
\end{figure}

At first, we investigate the impact of increasing network complexity on the privacy enhancing properties of PRECODE when attacking with IGA.
Fig.~\ref{fig:ssim_e1} displays the reconstruction quality as measured by SSIM when training the SMLP, DMLP, and LetNet architectures on CIFAR-10 with and without our proposed extension.
Whereas IGA on the baseline MLPs allowed for almost identical reconstructions with SSIM values close to $1$, the reconstructions showed negligible similarity after extension with PRECODE.
Also the results using LeNet with PRECODE display a massive reduction in SSIM, although SSIM of the baseline models was already reduced compared to the MLPs.

Table~\ref{tab:net_complexity} summarizes the results obtained on CIFAR-10.
Note that the reported accuracies might not be optimal due to potentially suboptimal hyperparameter choices. 
\begin{table}
\centering
\caption{Model performance and attack metrics of SMLP, DMLP, and LeNet models trained on CIFAR-10 without and with PRECODE (``Ours'').
Arrows indicate direction of improvement.
Bold formatting highlights best results.
}
\label{tab:net_complexity}
\resizebox{\linewidth}{!}{%
\begin{tabular}{lr | cc | cc | cc}
\toprule
Model && \multicolumn{2}{c}{SMLP} & \multicolumn{2}{|c}{DMLP} & \multicolumn{2}{|c}{LeNet} \\
Defense &&    Baseline &         Ours &    Baseline &         Ours &  Baseline &        Ours \\
\midrule\midrule
$\text{Acc}_\text{train}$ [\%] &$\uparrow$ &       \textbf{95.91} &       92.41 &       \textbf{98.51} &       98.22 &     \textbf{75.01} &      72.01 \\
$\text{Acc}_\text{test}$ [\%] &$\uparrow$  &       \textbf{53.07} &       51.75 &       \textbf{54.92} &       54.16 &     \textbf{61.19} &      58.99 \\
MSE &$\uparrow$            &        0.0 &       \textbf{3.95} &        0.0 &        \textbf{4.62} &      0.49 &       \textbf{2.17} \\
PSNR &$\downarrow$         &       44.13 &        \textbf{6.18} &       44.26 &        \textbf{5.51} &     15.72 &       \textbf{8.84} \\
SSIM &$\downarrow$         &        0.99 &        \textbf{0.03} &        0.99 &        \textbf{0.01} &      0.55 &       \textbf{0.1} \\
ASR  [\%] &$\downarrow$          &      100.0 &        \textbf{0.0} &      100.0 &        \textbf{0.0} &     42.45 &       \textbf{0.0} \\
\bottomrule
\end{tabular}
}
\end{table}

As can be seen, PRECODE had small impact on the accuracy, \ie, train and test accuracy slightly decreased.
Similar to the findings in~\cite{geiping2020inverting}, we observe no impact of the network depth on the reconstruction results.


\subsection{Defense Mechanisms}
\label{sec:exp_def}
\begin{table*}[!tb]
\centering
\caption{Performance and attack metrics of the DMLP and LeNet models trained on CIFAR-10 with and without the different defense mechanisms. 
Arrows indicate direction of improvement.
Bold formatting highlights best results.}
\label{tab:cif10}
\resizebox{.8\linewidth}{!}{%
\begin{tabular}{l r| c|cccc | c|cccc}
\toprule
Model && \multicolumn{5}{c}{DMLP} & \multicolumn{5}{|c}{LeNet} \\
Defense        && Baseline &  NG-2  &   NG-3 &     GC &   Ours  & Baseline &    NG-2 &   NG-3 &    GC &    Ours \\
\midrule\midrule
$\text{Acc}_\text{train}$ [\%] & $\uparrow$ &    \textbf{98.51} &  43.83 &  92.36 &  10.15 &  98.22 &    \textbf{75.01} &  56.88 &  73.55 &  22.29 &  72.01 \\
$\text{Acc}_\text{test}$ [\%] & $\uparrow$ &    \textbf{54.92} &  42.55 &  51.36 &  10.04 &  54.16 &    61.19 &  54.36 &  \textbf{62.13} &  23.40 &  58.99 \\
MSE            & $\uparrow$ &     0.0 &   2.5 &   0.17 &   1.13 &  \textbf{4.62} &     0.49 &   1.59 &   0.61 &   1.9 &   \textbf{2.17} \\
PSNR           &$\downarrow$&    44.26 &   8.11 &  18.64 &  12.09 &  \textbf{5.51} &    15.72 &  10.2 &  14.73 &   9.36 &   \textbf{8.84} \\
SSIM           &$\downarrow$&     0.99 &   0.03 &   0.62 &   0.42 &  \textbf{0.01} &     0.55 &   0.22 &   0.49 &   0.18 &   \textbf{0.1} \\
ASR [\%]       &$\downarrow$&   100.0 &   \textbf{0.0} &  63.80 &   3.12 &  \textbf{0.0} &    42.45 &   \textbf{0.0} &  30.47 &   \textbf{0.0} &   \textbf{0.0} \\
\bottomrule
\end{tabular}
}
\end{table*}

Next, we compare PRECODE to other state of the art defense mechanisms, \ie, \textit{noisy gradients} and \textit{gradient compression} by pruning.
For noisy gradients, we add Gaussian noise with zero mean and standard deviation $\sigma$.
We consider two noise levels: $\sigma=10^{-2}$ (NG-2) and $\sigma=10^{-3}$ (NG-3).
Wei \etal report Laplace noise to result in less privacy protection compared to Gaussian noise while having more negative impact on model accuracy~\cite{wei2020framework}.
For this reason, we did not consider Laplace noise in our experiments.
For gradient compression, referred to as GC in figures and tables, we prune $10\%$ of the gradient values with smallest magnitudes to zero as described in~\cite{zhu2019deep}.

\begin{figure}[!tb]
	\begin{center}
		\includegraphics*[width=1\linewidth]{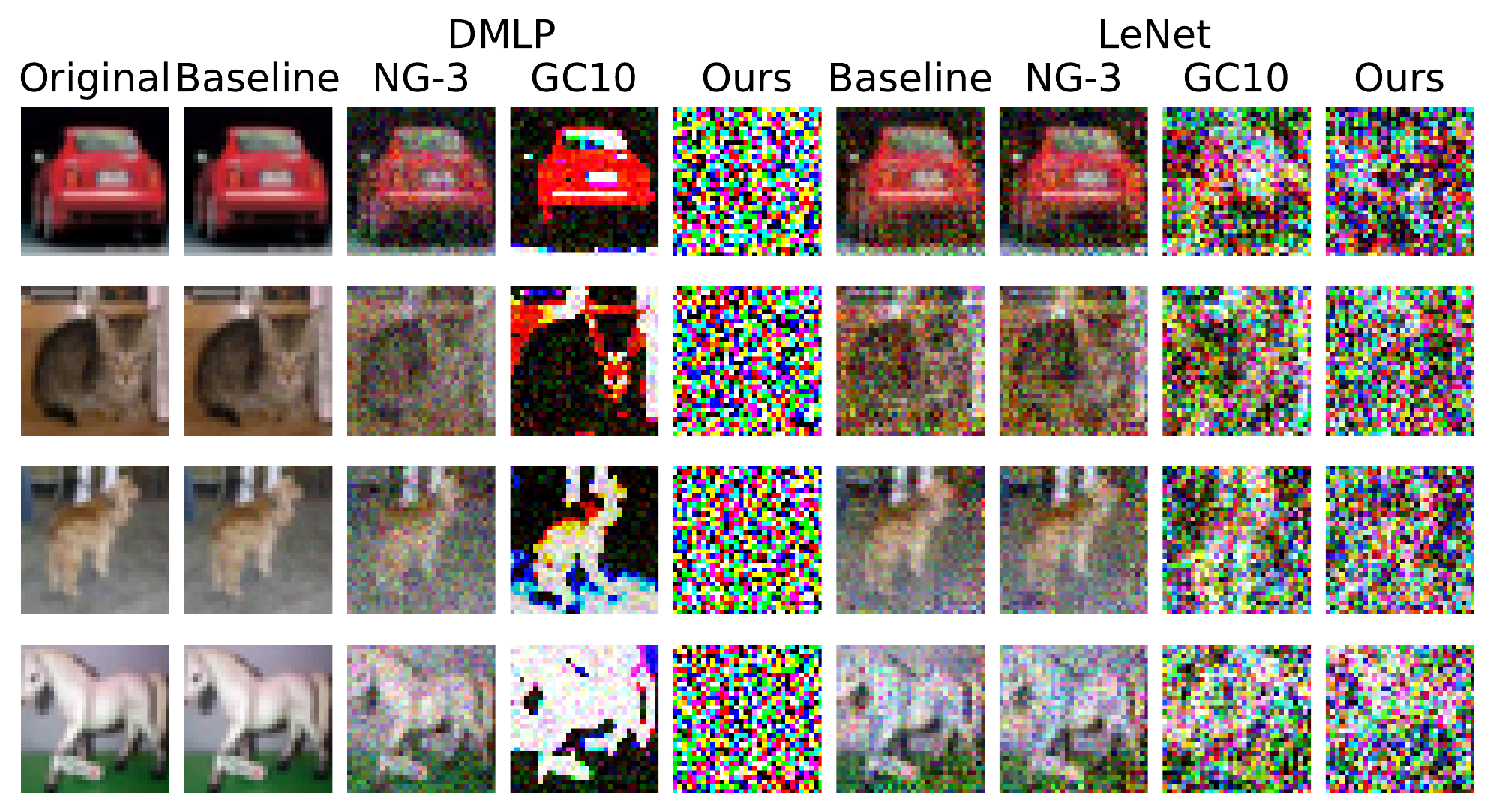}
        \setlength{\belowcaptionskip}{-10pt}
        \vspace{-15pt}
		\caption{Exemplary reconstruction results on the CIFAR-10 dataset for the baseline DMLP and LeNet models and different defense mechanisms.
        More reconstruction examples can be found in the \textit{supplementary material}.}
		\label{fig:cif10rec}
	\end{center}
\end{figure}

Fig.~\ref{fig:cif10rec} displays exemplary reconstruction results on the CIFAR-10 dataset.
As can be seen for both baseline models, the training data can be recognizably reconstructed, whereas extension with PRECODE prevents successful reconstruction.
Also reconstructions for models using NG-3 and GC show notably reduced similarity to the original images.
Yet, for most reconstructed images basic patterns and objects are still recognizable.

The metrics summarized in Table~\ref{tab:cif10} also reflect this behaviour: both NG-3 and GC show increasing MSE and decreasing PSNR, SSIM, and ASR.
However, PRECODE clearly outperforms both noisy gradients and gradient compression regarding privacy leakage prevention.
In addition, optimization using GC suffered from severe training instabilities that caused unacceptable performance drop.
Even with reduced compression rates, we were not able to train models with sufficient accuracy and therefore decided to exclude GC from further experiments.
Almost the same observation has been made using noisy gradients with increased noise levels, \ie, $\sigma=10^{-2}$ (NG-2).
NG-2 successfully prevents privacy leakage for both architectures but causes a notable drop in model accuracy.
Moderate noise levels, \ie, $\sigma = 10^{-3}$ (NG-3), cause less drop in model performance but allowed for larger SSIM.
For privacy preservation that is on par with PRECODE, noise levels greater than $10^{-2}$ would be required. 
However, this renders models practically irrelevant due to overall bad performance.
PRECODE on the other hand barely impacts model performance.

\begin{figure}[!tb]
	\begin{center}
		\includegraphics*[width=1\linewidth]{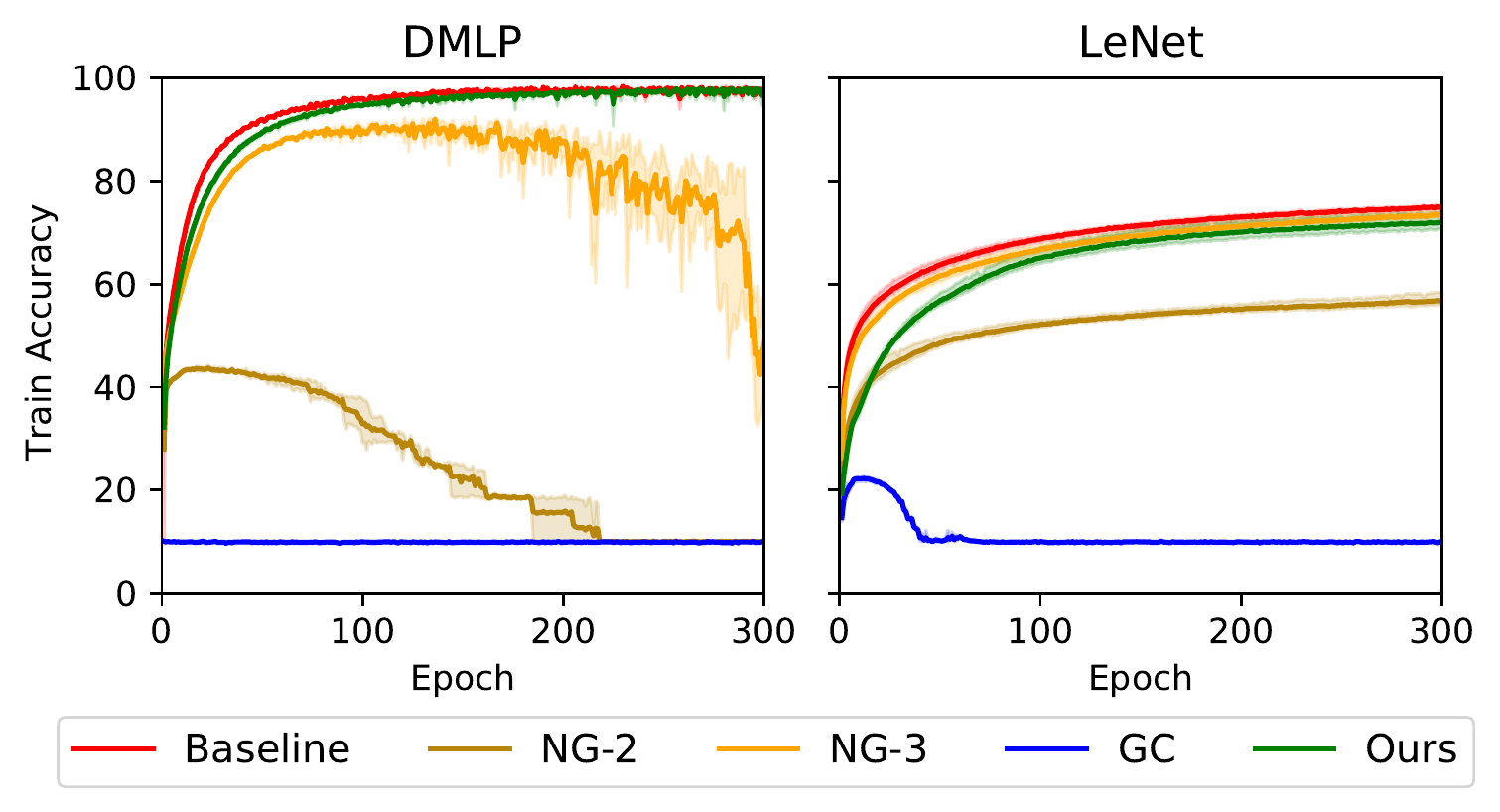}
        \setlength{\belowcaptionskip}{-10pt}
        \vspace{-15pt}
		\caption{Train accuracy on the CIFAR-10 dataset. Line colors define the baseline model and defense mechanisms.}
		\label{fig:cif10_trn_accuracies}
	\end{center}
\end{figure}

Fig.~\ref{fig:cif10_trn_accuracies} displays the learning curves in terms of training accuracy for the models trained on CIFAR-10 with and without defense mechanisms.
Whereas PRECODE shows no impact on the training process, severe training instabilities and reduced model performance can be observed for the DMLP with NG and GC.
NG-3 is the only approach that does not increase convergence time and decrease accuracy for LeNet.
However, PRECODE generally provides better protection against IGA, and achieves higher accuracy and faster convergence compared to NG-2 and GC.

\subsection{Different Datasets}
\label{sec:exp_data}
To extend the empirical evaluation to further datasets, we also consider the CIFAR-100 and Medical MNIST dataset.
\begin{figure}
\setlength{\belowcaptionskip}{0pt}
\captionof{table}{Model performance and attack metrics of the DMLP and LeNet models trained on (a) CIFAR-100, (b) Medical MNIST, with and without the different defense mechanisms. 
Arrows indicate direction of improvement.
Bold formatting highlights best results.}
    \label{tab:datasets}
    \centering
    \begin{subfigure}{\linewidth}
\label{tab:cif100}
\resizebox{\linewidth}{!}{%
\begin{tabular}{l r| c|ccc | c|ccc}
\toprule
Model && \multicolumn{4}{c}{DMLP} & \multicolumn{4}{|c}{LeNet} \\
Defense        && Baseline &  NG-2  &   NG-3  &   Ours            & Baseline &    NG-2 &   NG-3  &    Ours \\
\midrule\midrule
$\text{Acc}_\text{train}$ [\%]&$\uparrow$ &    61.85 &  14.55 &  64.78 &  \textbf{80.87} &    \textbf{63.73} &  21.45 &  48.23 &  54.42 \\
$\text{Acc}_\text{test}$ [\%]&$\uparrow$  &    \textbf{22.66} &  13.44 &  21.56 &  21.87 &    28.92 &  19.85 &  \textbf{30.22} &  27.37 \\
MSE &$\uparrow$            &     0.00 &   2.32 &   0.15 &   \textbf{4.14} &     0.81 &   1.35 &   0.76 &   \textbf{1.94} \\
PSNR &$\downarrow$         &    40.23 &   7.99 &  18.63 &   \textbf{5.55} &    12.69 &  10.39 &  13.17 &   \textbf{8.78} \\
SSIM &$\downarrow$         &     0.98 &   0.03 &   0.58 &   \textbf{0.01} &     0.33 &   0.20 &   0.36 &   \textbf{0.10} \\
ASR [\%]    &$\downarrow$  &   100.00 &   \textbf{0.00} &  45.05 &   \textbf{0.00} &     2.60 &   \textbf{0.00} &   8.33 &   \textbf{0.00} \\
\bottomrule
\end{tabular}
}
\caption{Result on CIFAR-100.}
    \end{subfigure}\\
    \vskip4pt
    \begin{subfigure}{\linewidth}
\label{tab:cpl}
\resizebox{\linewidth}{!}{%
\begin{tabular}{l r| c|ccc | c|ccc}
\toprule
Model && \multicolumn{4}{c}{DMLP} & \multicolumn{4}{|c}{LeNet} \\
Defense        && Baseline &  NG-2  &   NG-3  &   Ours                      & Baseline &    NG-2 &   NG-3  &    Ours \\
\midrule\midrule
$\text{Acc}_\text{train}$ [\%] &$\uparrow$ &    99.98 &  \textbf{100.00} &  99.96 &  99.89 &   \textbf{100.00} &  99.88 &  \textbf{100.00} &  \textbf{100.00} \\
$\text{Acc}_\text{test}$ [\%] &$\uparrow$  &    99.85 &   \textbf{99.88} &  99.82 &  99.80 &    \textbf{99.96} &  99.81 &   99.92 &   99.95 \\
MSE &$\uparrow$            &     0.00 &    2.86 &   0.17 &   \textbf{5.29} &     0.81 &   1.69 &    1.27 &    \textbf{2.35} \\
PSNR &$\downarrow$         &    38.93 &    6.57 &  16.15 &   \textbf{4.30} &    12.01 &   9.07 &   11.50 &    \textbf{7.55} \\
SSIM &$\downarrow$         &     0.94 &    0.03 &   0.43 &   \textbf{0.00} &     0.30 &   0.15 &    0.26 &    \textbf{0.09} \\
ASR [\%] &$\downarrow$      &    99.22 &    \textbf{0.00} &  10.94 &   \textbf{0.00} &    12.24 &   \textbf{0.00} &   12.24 &    \textbf{0.00} \\
\bottomrule
\end{tabular}
}
\caption{Results on Medical MNIST.}
    \end{subfigure}
\end{figure}
Results on CIFAR-100 are summarized in Table~\ref{tab:datasets}(a).
Similar to CIFAR-10, both PRECODE and NG-3 had marginal impact on the test accuracy, whereas increased noise levels of NG-2 resulted in a drop of 10\%.
Comparing privacy leakage metrics, extending the models with PRECODE resulted in largest MSE and smallest PSNR, SSIM, and ASR.

The results obtained on Medical MNIST, summarized in Table~\ref{tab:datasets}(b), also confirm these observations.
Compared to CIFAR-10 and CIFAR-100, classification on Medical MNIST generally appears a simpler task as all models achieved $100\%$ accuracy within the first few epochs.

Regarding model training, PRECODE showed the most stable optimization process and fastest model convergence across the privacy-preserving models.
The learning curves for both datasets can be found in the \textit{supplementary material}.
For some runs, NG-3 showed faster convergence and increased model performance but also increased privacy leakage compared to NG-2 and PRECODE.


\subsection{Different Attacks}
\label{sec:exp_att}
To show the effectiveness of PRECODE against other attacks we conduct a series of experiments using the CPL attack~\cite{wei2020framework}.
Therefore, we adjusted the reconstruction loss function by replacing cosine distance with Euclidean distance as described in section~\ref{sec:sota_rec_att}.
When using the originally reported L-BFGS optimizer with learning rate $1$~\cite{wei2020framework}, we observed notably increased optimization time and comparably bad reconstruction results even for the baseline models.
For this reason, we exchanged L-BFGS with Adam optimizer using parameters as described in section~\ref{sec:exp_setup}.

Table~\ref{tab:cpl} shows the results obtained on CIFAR-10.
\begin{table}
\centering
\caption{Attack metrics of the DMLP and LeNet models for the CPL attack on CIFAR-10 with and without the different defense mechanisms. 
Arrows indicate direction of improvement.
Bold formatting highlights best results.
}
\label{tab:cpl}
\resizebox{\linewidth}{!}{%
\begin{tabular}{l r| c|ccc | c|ccc}
\toprule
Model && \multicolumn{4}{c}{DMLP} & \multicolumn{4}{|c}{LeNet} \\
Defense        && Baseline &  NG-2  &   NG-3  &   Ours           & Baseline &    NG-2 &   NG-3  &    Ours \\
\midrule\midrule
MSE &$\uparrow$    &     0.0 &  3.01 &   0.17 &  \textbf{4.07} &     0.67 &   1.52 &   0.72 &  \textbf{2.51} \\
PSNR &$\downarrow$ &    40.53 &  7.31 &  18.57 &  \textbf{6.05} &    14.53 &  10.33 &  14.15 &  \textbf{8.12} \\
SSIM &$\downarrow$ &     0.99 &  0.03 &   0.62 &  \textbf{0.01} &     0.47 &   0.22 &   0.45 &  \textbf{0.06} \\
ASR [\%]&$\downarrow$  &   100.0 &  \textbf{0.0} &  64.58 &  \textbf{0.0} &    29.17 &   \textbf{0.0} &  27.86 &  \textbf{0.0} \\
\bottomrule
\end{tabular}
}
\end{table}

The overall behavior is similar to the results obtained using IGA.
However, compared to IGA (cf.~Table~\ref{tab:cif10}), we found CPL to generally achieve slightly worse reconstructions.


\subsection{Batchsizes and Training Progress}
\label{sec:exp_prog}
Last, we investigate privacy leakage prevention by PRECODE over different model training states and batch sizes.
All models are attacked after random initialization at the beginning of the training.
In addition, we attack the models after $1$ and $50$ epochs of training.
We specifically choose these checkpoints to investigate privacy leakage after each model was trained exactly once on each sample of the training set, as well as when the models are close to convergence (cf. Fig.~\ref{fig:cif10_trn_accuracies}).
\begin{figure}[!tb]
	\begin{center}
		\includegraphics*[width=1\linewidth]{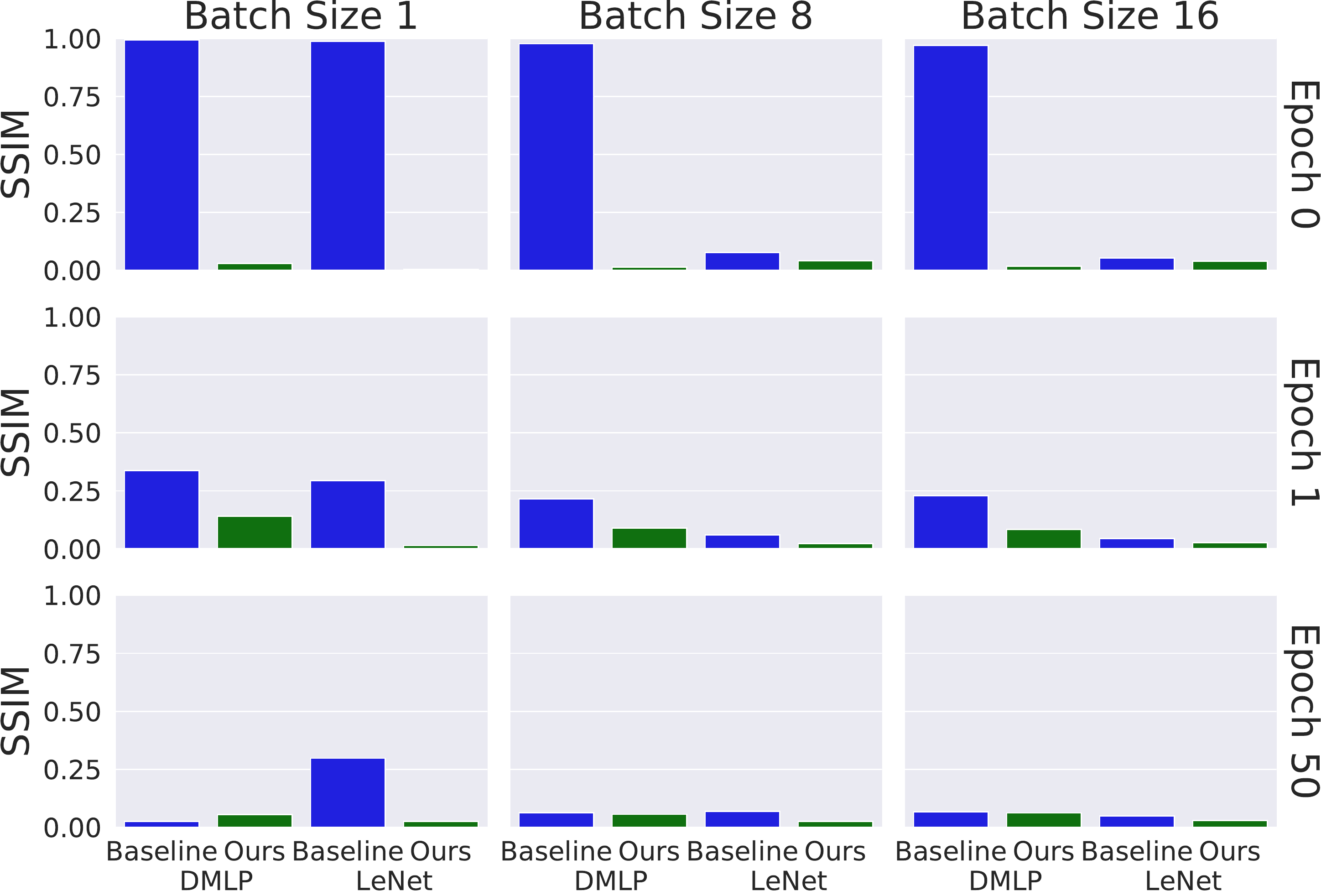}
		\caption{SSIM of reconstructed images by IGA on the DMLP, and LeNet models with and without PRECODE for varying victim batch sizes and model training states.}
		\label{fig:ckpt_bs_ssim}
	\end{center}
\end{figure}
Regarding batch sizes, we consider a batch size of $1$ as baseline, and further investigate models attacked with batches of size $8$ and $16$.
All attacks are performed using the same \textit{victim dataset} as described in section~\ref{sec:exp_setup}.
Due to the long computational time needed to execute all reconstruction attacks with these batch sizes, we only consider the baseline models and their extension with PRECODE.

The evaluated reconstruction results are displayed in Fig.~\ref{fig:ckpt_bs_ssim} in terms of SSIM.
During training of the DMLP the victim dataset can be almost entirely reconstructed with SSIM close to $1$ independent of the batch size.
For LeNet however, increasing batch size caused decreasing SSIM.
It can also be observed that SSIM decreases with progressing training state.
This can be attributed to lower gradient magnitudes induced by reduced loss as model predictions become more accurate.

Extending the models architectures with PRECODE almost entirely results in decreased SSIM compared to the baseline models.
With progressing training using PRECODE, a slight increase in SSIM can be observed.
However, all SSIM values remain below $0.15$, indicating that no recognizable visual structures can be reconstructed.


\section{Discussion}
\label{sec:discussion}

\subsection{Key Takeaways}

Confirming findings of earlier studies~\cite{geiping2020inverting, wei2020framework, zhao2020idlg, zhu2019deep}, all considered models are prone to privacy leakage by their gradients.
Especially MLPs, \ie, models solely constituted of fully connected layers, allow near-identical reconstructions of original training data.
To enhance privacy, gradients can be perturbed by noise addition or compression.
However, in order to successfully defend against gradient inversion attacks, levels of gradient perturbation that severely impact the optimization process and model performance are required.
We conclude that there is an inherent trade-off between model performance and privacy if gradient perturbation methods are used.

Our proposed model extension PRECODE avoids such trade-off.
Comparative evaluation using different baseline models and datasets revealed remarkable performance in preventing privacy leakage by IGA and CPL attacks, confirming PRECODE to meet the \textbf{privacy enhancement} requirement.
By design, PRECODE is a \textbf{generic extension} applicable to arbitrary model architectures by \textbf{direct integration} into the original model and joint optimization.
Yet, PRECODE is a trainable module, \ie, the computational costs increase as the number of trainable model parameters are increased.
Realized by fully connected layers, the amount of parameters added by PRECODE grows linearly with the size of the latent features to be stochastically encoded as well as the size of the variational bottleneck. 
Compared to, \eg, cryptographic methods, this renders negligible increase in computational costs.
More importantly, PRECODE showed no negative effect on the convergence time.
Contrary to gradient perturbation at privacy enhancing levels, PRECODE entirely remains within the \textbf{model performance} requirement.

\subsection{Limitations}
Our study is limited as only optimization based gradient inversion attacks are considered.
There is no guarantee that other, \eg, analytical attacks \cite{pan2020theory, zhu2021r}, are prevented with same efficiency.
However, such attacks would need to find sophisticated ways to account for the stochastic sampling, analogous to the reparameterization trick used for backpropagation during variational modeling.
In addition, we closely followed related work and investigated only deep learning tasks showcasing very successful attacks on comparably small models, \ie, image classification using MLPs and LeNet.
Whereas our proposed extension should also be applicable to other domains such as natural language processing, we currently provide no empirical proof.
Regarding model complexity, Geiping \etal~\cite{geiping2020inverting} showed that sufficient parameter tuning and restart attempts even allow to reconstruct images from large batches and more complex models such as residual neural networks~\cite{he2016deep}.
Each attack optimization step then takes orders of magnitudes more time, which prevented systematic evaluation within this study.


\section{Conclusion}
\label{sec:conc}
We introduced PRECODE -- PRivacy EnhanCing mODulEs -- to defend deep learning models against privacy leakage by gradients exchanged during collaborative learning.
Inspired by variational modeling, we realized PRECODE as variational bottleneck prior to the output layer of original models.
We systematically evaluated our extension using fully connected and convolutional architectures on three image classification datasets and two state of the art gradient inversion attacks.
To prevent privacy leakage, commonly used gradient perturbation methods require perturbation levels that cause unacceptable drop in model performance.
Compared to these methods, PRECODE showed consistently better prevention of privacy leakage, while having almost no impact on model performance.
We conclude that our proposed extension represents a promising alternative to commonly used defense mechanisms.
In future work we plan to evaluate other model architectures, datasets, and tasks, for deeper investigation of the generic applicability of our approach.

\section{Acknowledgments}
\label{sec:ack}
We are funded by the Thuringian Ministry for Economic Affairs, Science and Digital Society (Grant: 5575/10-3).

{\small
\bibliographystyle{ieee_fullname}
\bibliography{bibliography}

\begin{thebibliography}{10}\itemsep=-1pt

\bibitem{abadi2016deep}
Martin Abadi, Andy Chu, Ian Goodfellow, H~Brendan McMahan, Ilya Mironov, Kunal
  Talwar, and Li Zhang.
\newblock Deep learning with differential privacy.
\newblock In {\em Proceedings of the 2016 ACM SIGSAC Conference on Computer and
  Communications Security}, pages 308--318, 2016.

\bibitem{alemi2016deep}
Alexander~A Alemi, Ian Fischer, Joshua~V Dillon, and Kevin Murphy.
\newblock Deep variational information bottleneck.
\newblock {\em arXiv preprint arXiv:1612.00410}, 2016.

\bibitem{aono2017privacy}
Yoshinori Aono, Takuya Hayashi, Lihua Wang, Shiho Moriai, et~al.
\newblock Privacy-preserving deep learning via additively homomorphic
  encryption.
\newblock {\em IEEE Transactions on Information Forensics and Security},
  13(5):1333--1345, 2017.

\bibitem{apolanco2017med}
A.~P.~Lozano (apolanco3225).
\newblock Medical mnist classification.
\newblock \url{https://github.com/apolanco3225/Medical-MNIST-Classification},
  2017.

\bibitem{bonawitz2017practical}
Keith Bonawitz, Vladimir Ivanov, Ben Kreuter, Antonio Marcedone, H~Brendan
  McMahan, Sarvar Patel, Daniel Ramage, Aaron Segal, and Karn Seth.
\newblock Practical secure aggregation for privacy-preserving machine learning.
\newblock In {\em Proceedings of the 2017 ACM SIGSAC Conference on Computer and
  Communications Security}, pages 1175--1191, 2017.

\bibitem{duan2020self}
Moming Duan, Duo Liu, Xianzhang Chen, Renping Liu, Yujuan Tan, and Liang Liang.
\newblock Self-balancing federated learning with global imbalanced data in
  mobile systems.
\newblock {\em IEEE Transactions on Parallel and Distributed Systems},
  32(1):59--71, 2020.

\bibitem{dwork2014algorithmic}
Cynthia Dwork, Aaron Roth, et~al.
\newblock The algorithmic foundations of differential privacy.
\newblock {\em Foundations and Trends in Theoretical Computer Science},
  9(3-4):211--407, 2014.

\bibitem{geiping2020inverting}
Jonas Geiping, Hartmut Bauermeister, Hannah Dr\"{o}ge, and Michael Moeller.
\newblock Inverting gradients - how easy is it to break privacy in federated
  learning?
\newblock In H. Larochelle, M. Ranzato, R. Hadsell, M.~F. Balcan, and H. Lin,
  editors, {\em Advances in Neural Information Processing Systems}, volume~33,
  pages 16937--16947. Curran Associates, Inc., 2020.

\bibitem{he2016deep}
Kaiming He, Xiangyu Zhang, Shaoqing Ren, and Jian Sun.
\newblock Deep residual learning for image recognition.
\newblock In {\em Proceedings of the IEEE conference on computer vision and
  pattern recognition}, pages 770--778, 2016.

\bibitem{jayaraman2019evaluating}
Bargav Jayaraman and David Evans.
\newblock Evaluating differentially private machine learning in practice.
\newblock In {\em 28th $\{$USENIX$\}$ Security Symposium ($\{$USENIX$\}$
  Security 19)}, pages 1895--1912, 2019.

\bibitem{kairouz2019advances}
Peter Kairouz, H~Brendan McMahan, Brendan Avent, Aur{\'e}lien Bellet, Mehdi
  Bennis, Arjun~Nitin Bhagoji, Keith Bonawitz, Zachary Charles, Graham Cormode,
  Rachel Cummings, et~al.
\newblock Advances and open problems in federated learning.
\newblock {\em arXiv preprint arXiv:1912.04977}, 2019.

\bibitem{kingma2015adam}
Diederik~P Kingma and Jimmy Ba.
\newblock Adam: A method for stochastic optimization.
\newblock In {\em In International Conference on Learning Representations
  (ICLR)}, May 2015.

\bibitem{kingma2013auto}
Diederik~P Kingma and Max Welling.
\newblock Auto-encoding variational bayes.
\newblock {\em arXiv preprint arXiv:1312.6114}, 2013.

\bibitem{konevcny2016federated}
Jakub Kone{\v{c}}n{\`y}, H~Brendan McMahan, Felix~X Yu, Peter Richt{\'a}rik,
  Ananda~Theertha Suresh, and Dave Bacon.
\newblock Federated learning: Strategies for improving communication
  efficiency.
\newblock {\em arXiv preprint arXiv:1610.05492}, 2016.

\bibitem{krizhevsky2009learning}
Alex Krizhevsky, Geoffrey Hinton, et~al.
\newblock Learning multiple layers of features from tiny images.
\newblock 2009.

\bibitem{lee2020tornadoaggregate}
Jin-woo Lee, Jaehoon Oh, Sungsu Lim, Se-Young Yun, and Jae-Gil Lee.
\newblock Tornadoaggregate: Accurate and scalable federated learning via the
  ring-based architecture.
\newblock {\em arXiv preprint arXiv:2012.03214}, 2020.

\bibitem{li2019privacy}
Wenqi Li, Fausto Milletar{\`\i}, Daguang Xu, Nicola Rieke, Jonny Hancox, Wentao
  Zhu, Maximilian Baust, Yan Cheng, S{\'e}bastien Ourselin, M~Jorge Cardoso,
  et~al.
\newblock Privacy-preserving federated brain tumour segmentation.
\newblock In {\em International Workshop on Machine Learning in Medical
  Imaging}, pages 133--141. Springer, 2019.

\bibitem{lin2017deep}
Yujun Lin, Song Han, Huizi Mao, Yu Wang, and William~J Dally.
\newblock Deep gradient compression: Reducing the communication bandwidth for
  distributed training.
\newblock {\em arXiv preprint arXiv:1712.01887}, 2017.

\bibitem{liu1989limited}
Dong~C Liu and Jorge Nocedal.
\newblock On the limited memory bfgs method for large scale optimization.
\newblock {\em Mathematical programming}, 45(1):503--528, 1989.

\bibitem{mcmahan2017communication}
Brendan McMahan, Eider Moore, Daniel Ramage, Seth Hampson, and Blaise~Aguera y
  Arcas.
\newblock Communication-efficient learning of deep networks from decentralized
  data.
\newblock In {\em Artificial Intelligence and Statistics}, pages 1273--1282.
  PMLR, 2017.

\bibitem{mcmahan2018learning}
H~Brendan McMahan, Daniel Ramage, Kunal Talwar, and Li Zhang.
\newblock Learning differentially private recurrent language models.
\newblock In {\em International Conference on Learning Representations}, 2018.

\bibitem{pan2020theory}
Xudong Pan, Mi Zhang, Yifan Yan, Jiaming Zhu, and Min Yang.
\newblock Theory-oriented deep leakage from gradients via linear equation
  solver.
\newblock {\em arXiv preprint arXiv:2010.13356}, 2020.

\bibitem{roy2019braintorrent}
Abhijit~Guha Roy, Shayan Siddiqui, Sebastian P{\"o}lsterl, Nassir Navab, and
  Christian Wachinger.
\newblock Braintorrent: A peer-to-peer environment for decentralized federated
  learning.
\newblock {\em arXiv preprint arXiv:1905.06731}, 2019.

\bibitem{rudin1992nonlinear}
Leonid~I Rudin, Stanley Osher, and Emad Fatemi.
\newblock Nonlinear total variation based noise removal algorithms.
\newblock {\em Physica D: nonlinear phenomena}, 60(1-4):259--268, 1992.

\bibitem{sattler2019robust}
Felix Sattler, Simon Wiedemann, Klaus-Robert M{\"u}ller, and Wojciech Samek.
\newblock Robust and communication-efficient federated learning from non-iid
  data.
\newblock {\em IEEE transactions on neural networks and learning systems},
  31(9):3400--3413, 2019.

\bibitem{shokri2015privacy}
Reza Shokri and Vitaly Shmatikov.
\newblock Privacy-preserving deep learning.
\newblock In {\em Proceedings of the 22nd ACM SIGSAC conference on computer and
  communications security}, pages 1310--1321, 2015.

\bibitem{tagliavini2018transprecision}
Giuseppe Tagliavini, Stefan Mach, Davide Rossi, Andrea Marongiu, and Luca
  Benini.
\newblock A transprecision floating-point platform for ultra-low power
  computing.
\newblock In {\em 2018 Design, Automation \& Test in Europe Conference \&
  Exhibition (DATE)}, pages 1051--1056. IEEE, 2018.

\bibitem{wagner2018technical}
Isabel Wagner and David Eckhoff.
\newblock Technical privacy metrics: a systematic survey.
\newblock {\em ACM Computing Surveys (CSUR)}, 51(3):1--38, 2018.

\bibitem{wang2020sapag}
Yijue Wang, Jieren Deng, Dan Guo, Chenghong Wang, Xianrui Meng, Hang Liu,
  Caiwen Ding, and Sanguthevar Rajasekaran.
\newblock Sapag: A self-adaptive privacy attack from gradients.
\newblock {\em arXiv preprint arXiv:2009.06228}, 2020.

\bibitem{wang2004image}
Zhou Wang, Alan~C Bovik, Hamid~R Sheikh, and Eero~P Simoncelli.
\newblock Image quality assessment: from error visibility to structural
  similarity.
\newblock {\em IEEE transactions on image processing}, 13(4):600--612, 2004.

\bibitem{wei2020framework}
Wenqi Wei, Ling Liu, Margaret Loper, Ka-Ho Chow, Mehmet~Emre Gursoy, Stacey
  Truex, and Yanzhao Wu.
\newblock A framework for evaluating gradient leakage attacks in federated
  learning.
\newblock {\em arXiv preprint arXiv:2004.10397}, 2020.

\bibitem{zhao2020idlg}
Bo Zhao, Konda~Reddy Mopuri, and Hakan Bilen.
\newblock idlg: Improved deep leakage from gradients.
\newblock {\em arXiv preprint arXiv:2001.02610}, 2020.

\bibitem{zhu2021r}
Junyi Zhu and Matthew Blaschko.
\newblock R-gap: Recursive gradient attack on privacy.
\newblock {\em Proceedings ICLR 2021}, 2021.

\bibitem{zhu2019deep}
Ligeng Zhu, Zhijian Liu, and Song Han.
\newblock Deep leakage from gradients.
\newblock In {\em Advances in Neural Information Processing Systems}, pages
  14774--14784, 2019.

\end{thebibliography}
}

\clearpage
\begin{strip}  
\vspace{-30pt}
\begin{center}
      {\Large \bf Supplementary Material}
      \vspace*{12pt}
\end{center}     
\end{strip}    

\appendix

\section{Overview}
\begin{figure}[!hb]
	\begin{center}
		\includegraphics*[width=1\linewidth]{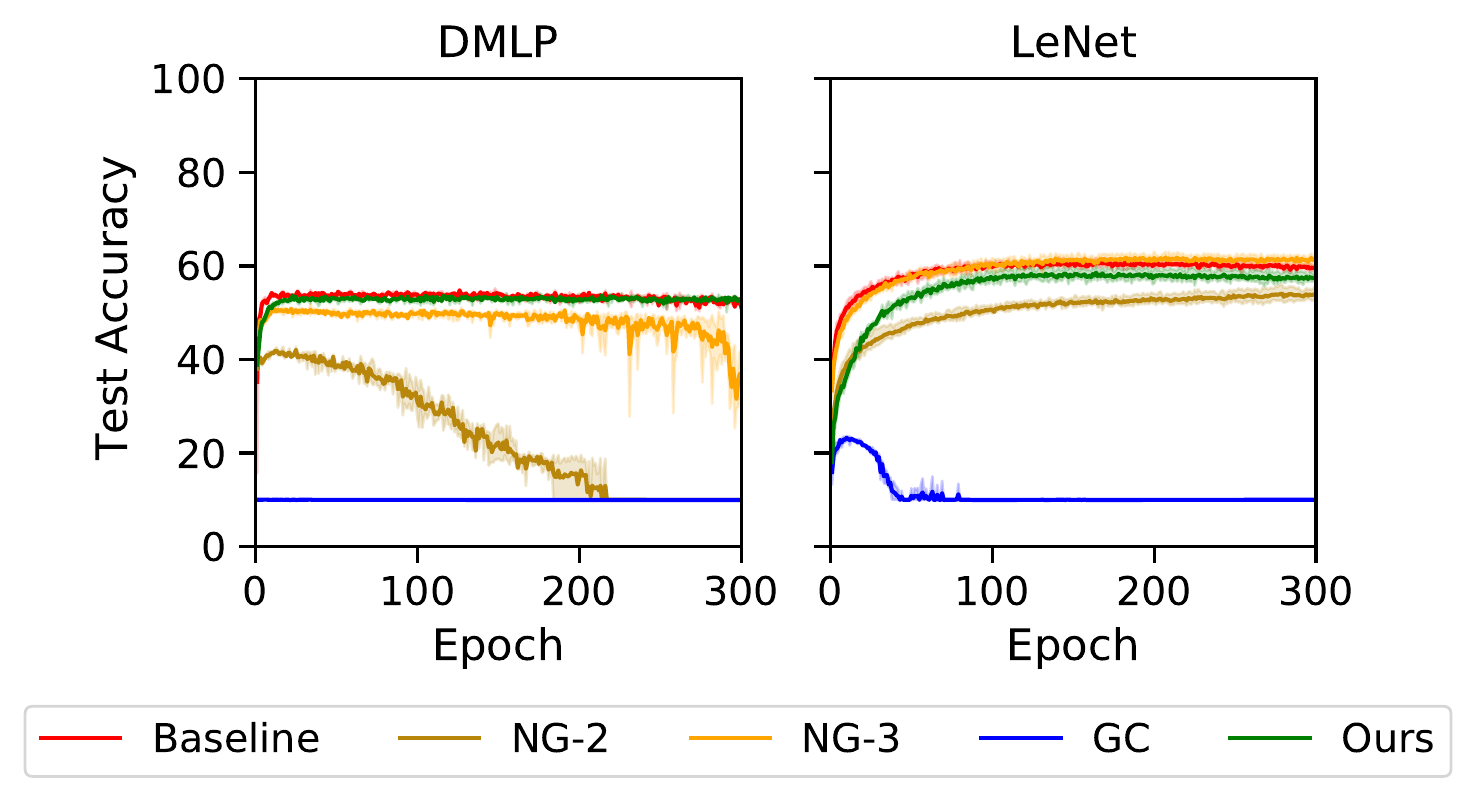}
		\caption{Test accuracy on the CIFAR-10 dataset. Line colors define the baseline model and defense mechanisms.}
		\label{fig:cif10_tst_accuracies}
	\end{center}
\end{figure}

\begin{figure}
     \centering
     \begin{subfigure}{\linewidth}
         \centering
         \includegraphics[width=\linewidth]{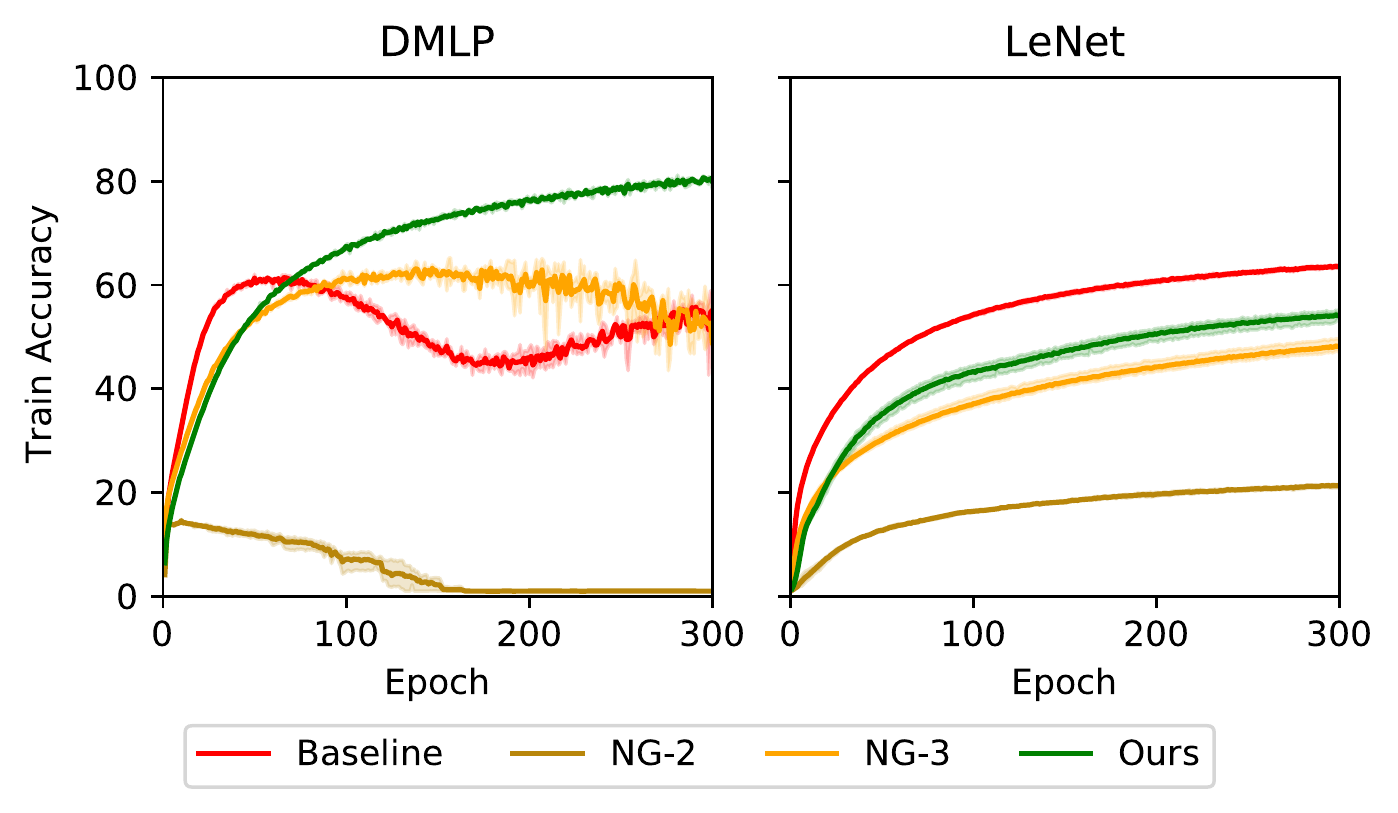}  
         \caption{}
     \end{subfigure}
     \begin{subfigure}{\linewidth}
         \centering
         \includegraphics[width=\linewidth]{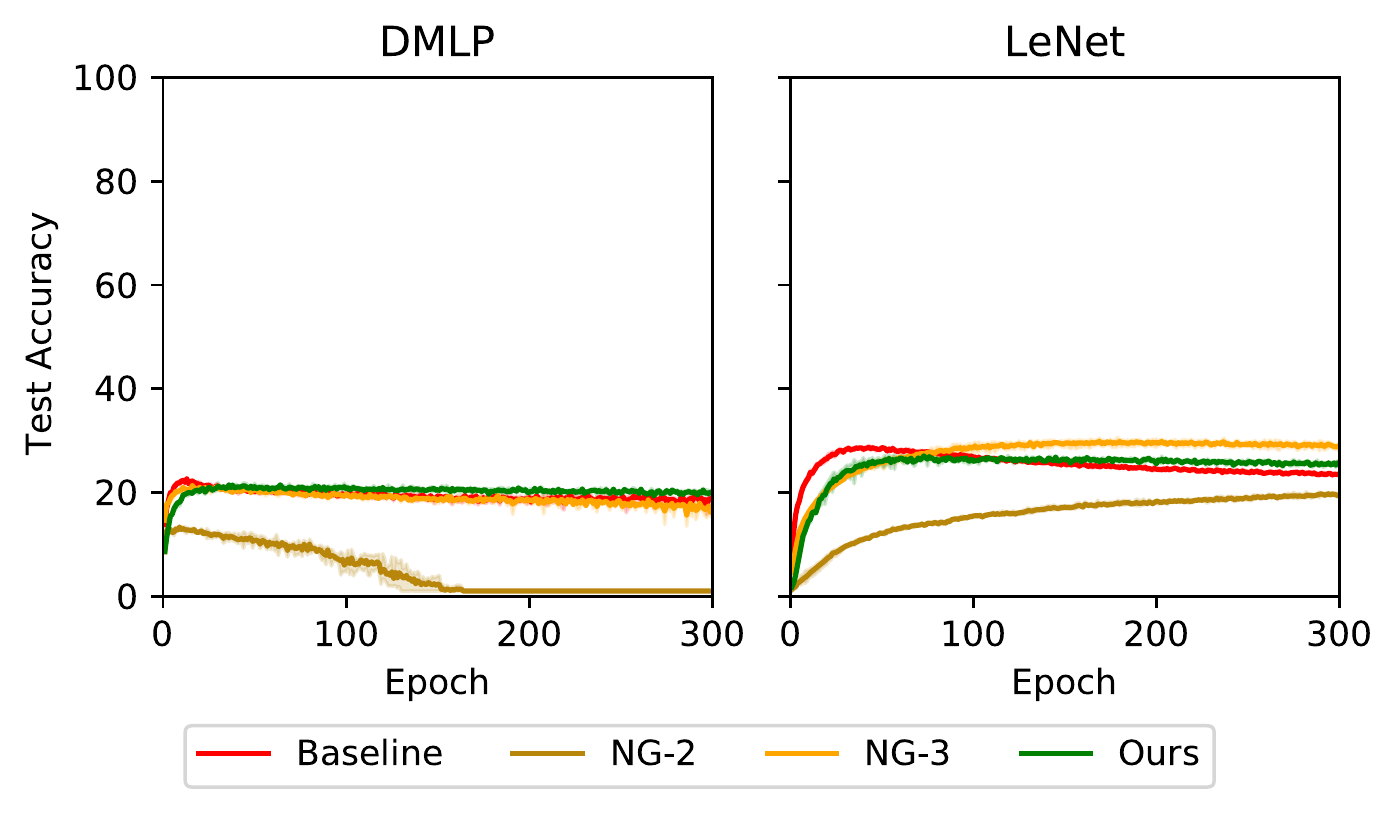}
         \caption{}
     \end{subfigure}
     \caption{Train accuracy (a) and test accuracy (b) on the CIFAR-100 dataset. Line colors define the baseline model and defense mechanisms.}
     \label{fig:cif100_accuracies}
\end{figure}

\begin{figure}
     \centering
     \begin{subfigure}{\linewidth}
         \centering
         \includegraphics[width=\linewidth]{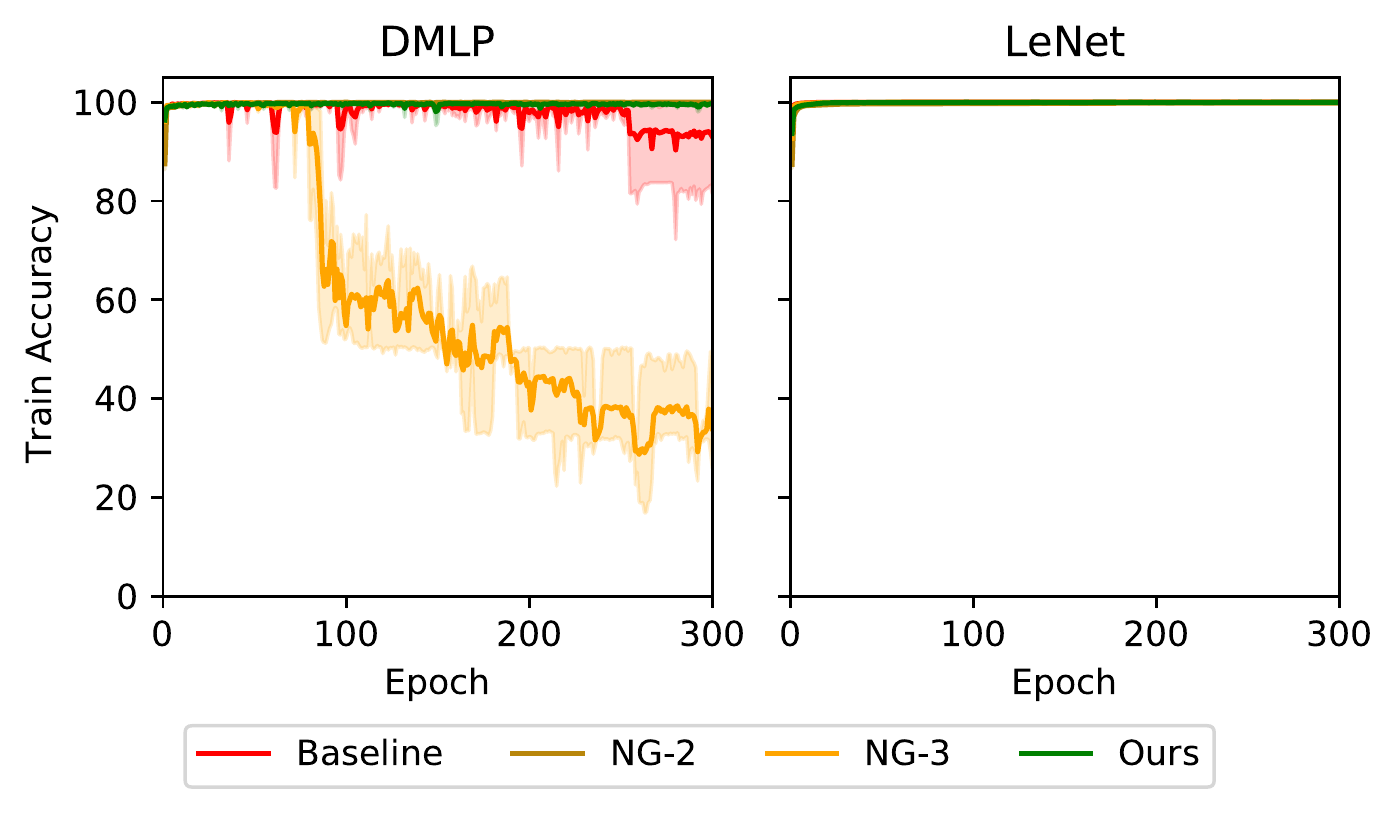}  
         \caption{}
     \end{subfigure}
     \begin{subfigure}{\linewidth}
         \centering
         \includegraphics[width=\linewidth]{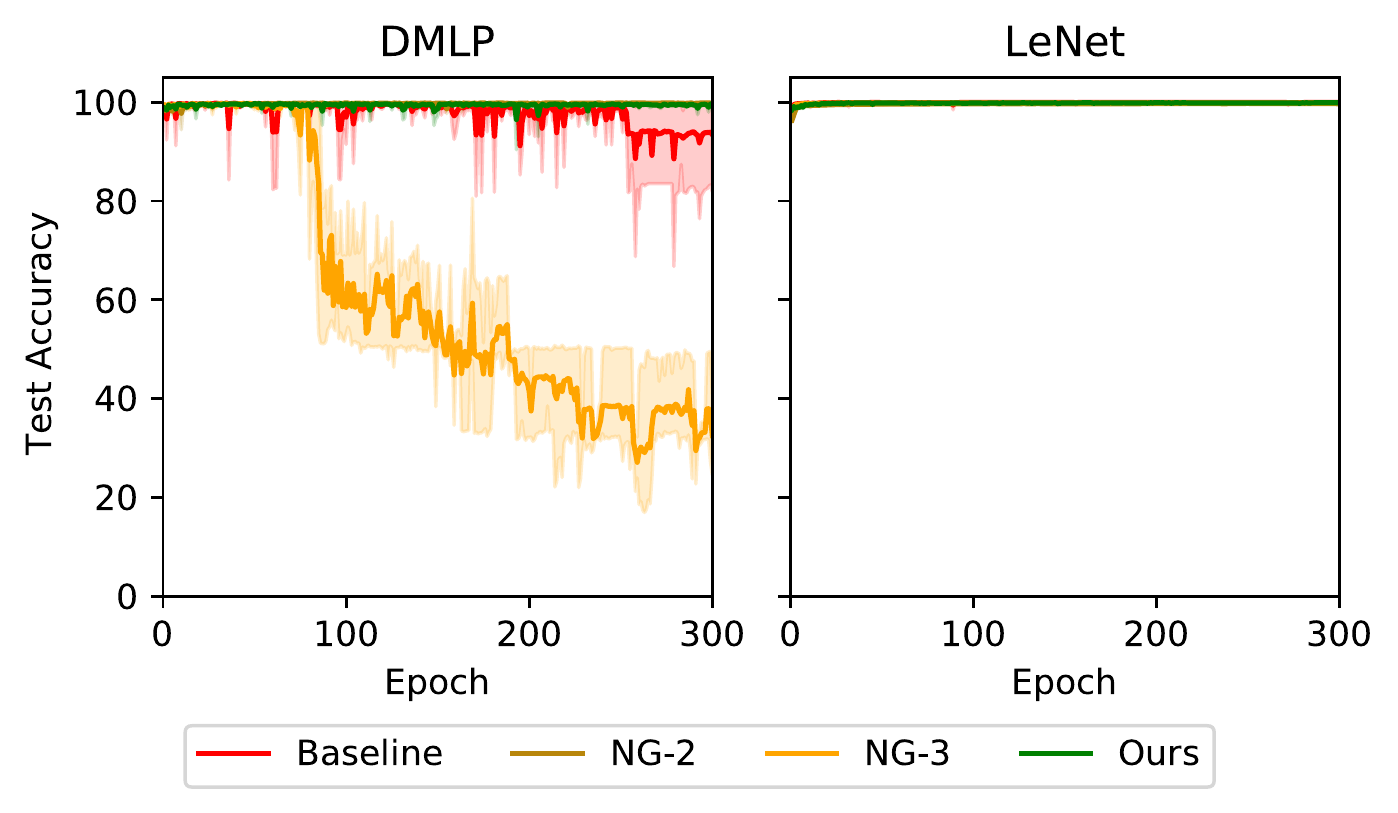}
         \caption{}
     \end{subfigure}
     \caption{Train accuracy (a) and test accuracy (b) on the Medical MNIST dataset. Line colors define the baseline model and defense mechanisms.}
     \label{fig:mm_accuracies}
\end{figure}

In sections~\ref{sec:ALF} and~\ref{sec:metr} we specify the loss functions used for the reconstruction of images in the conducted experiments as well as the applied metrics for evaluating the attacks.
Furthermore, the hyperparameters, positioning and quantity of PRECODE modules are discussed in section~\ref{sec:hyper}.
Finally we provide the accuracy curves that describe the models training progress (Fig.~\ref{fig:cif10_tst_accuracies}-\ref{fig:mm_accuracies}) and more exemplary reconstruction results for the CIFAR-10, CFIAR-100 and Medical MNIST datasets (Fig.~\ref{fig:cif10rec}-\ref{fig:mmrec}).

\section{Attack Loss Functions}
\label{sec:ALF}

The Client Privacy Leakage (CPL) attack~\cite{wei2020framework} extends the objective of minimizing the euclidean distance of the dummy gradient and the observed gradient as in DLG/iDLG by a label-based regularization term to stabilize the optimization.
\begin{equation}
    \label{eq:CPL}
    \begin{aligned}
        \arg \min_{x'} ||\nabla \mathcal{L}_\theta(F(x), y) - \nabla \mathcal{L}_\theta(F(x'),y)||^2 \\  + \alpha ||F(x'),y||^2,
    \end{aligned}
\end{equation}
where $\alpha$ tunes the impact of the regularization term on the optimization.

To perform the IGA attack described in~\cite{geiping2020inverting} the dummy image $x'$ is optimized for:
\begin{equation}
    \begin{aligned}
        \arg \min_{x'} 1 - \frac{\nabla \mathcal{L}_\theta(F(x), y) \cdot \nabla \mathcal{L}_\theta(F(x'),y)}{||\nabla \mathcal{L}_\theta(F(x), y)|| ||\nabla \mathcal{L}_\theta(F(x'),y)||} \\ + \alpha \mathrm{TV}(x').
    \end{aligned}
    \label{eq:GI}
\end{equation}
The total variation of the reconstructed image $\mathrm{TV}(x')$ is added as a simple image prior, where $\alpha$ determines its weight during optimization.
The output labels $y$ can be easily determined from the gradients as discussed before~\cite{zhao2020idlg}. 

\section{Metrics}
\label{sec:metr}
In our experiments MSE measures the pixelwise squared error as:
\begin{equation}
    \label{eq:mse}
    \text{MSE}(x, x') = \frac{1}{n_p}\sum^{n_p}_{i=1}(x_i-x'_i)^2, 
\end{equation}
where $n_p$ is the number of pixels in the images $x$ and $x'$.
Lower MSE values indicate a higher image similarity.
PSNR, which is commonly used to assess lossy compression schemes in video and imaging, is defined as:
\begin{equation}
    \label{eq:psnr}
    \text{PSNR}(x, x') = 10\log_{10}\left(\frac{\max(x)^2}{\text{MSE}(x, x')}\right)
\end{equation}
The higher the PSNR value the better the better the reconstruction.
SSIM uses a perception based model to measure the structural similarity between two images~\cite{wang2004image}:
\begin{equation}
    \label{eq:ssim}
    \text{SSIM}(x, x') = \frac{(2\mu_x\mu_{x'}+c_1)(2\sigma_{xx'}+c_2)}{(\mu^2_x+\mu^2_{x'}+c_1)(\sigma^2_x+\sigma^2_{x'}+c_2)},
\end{equation}
where $\mu_x$, $\sigma_{x}$ and $\mu_{x'}$, $\sigma_{x'}$ are the mean and variance values for $x$ and $x'$ respectively. 
$\sigma_{xx'}$ denotes the covariance of $x$ and $x'$, $c_1 = (k_1L)^2$ and $c_2 = (k_2L)^2$.
$k_1=0.01$ and $k_2=0.03$ are set by default and $L$ is the dynamic range of pixel values.
SSIM values closer to 1 indicate a higher similarity between images.

\section{PRECODE VB Hyperparameters and Positioning}
\label{sec:hyper}
We investigated the influence of the VB hyperparameters $k \in \lbrack 64, 128, 256, 512 \rbrack$ and $\beta \in \lbrack 10^{-1}, 10^{-2}, 10^{-3}, 10^{-4} \rbrack$ as well as the position and quantity of PRECODE modules in the model on the training process and quality of reconstructions.

If multiple bottlenecks are added to the model, the loss function is adjusted accordingly:
\begin{equation}
 \label{eq:multi_vb}
 \mathcal{L}(\hat{y}, y) = \mathcal{L}_F(\hat{y}, y) + \sum_{i=0}^{n} \mathcal{L}_{P_i},
\end{equation}
where 
\begin{equation}
 \label{eq:multi_vb}
 \mathcal{L}_{P_i} = \beta_i D_{KL}(\mathcal{N}(\mu_{B_i}, \sigma_{B_i}), \mathcal{N}(0, 1)),
\end{equation}
and $\beta_i$ tunes the weight of the $i$th PRECODE module $P_i$ with bottleneck layer $B_i$ on the overall loss function.

We found that the choice of $k$ had no notable impact on the training process, the final model performance and the quality of reconstructions.
If the choice for $\beta$ was too high (\ie $\beta = 10^{-1}$), the time for convergence increased and model performance decreased.
For the other $\beta$ values there was no further impact observed.
Again we found no notable impact on the quality of reconstructions.
We placed PRECODE after four different feature extracting layers in a MLP with four hidden layers and found that a placement too close to the input resulted in a slightly increased quality of image reconstructions.
Nevertheless the content of the training data was still unrecognizable and model performance was not impacted.
Placing more than one PRECODE module into the network resulted in a decrease of model performance but did not impact the quality of reconstructions.

\clearpage

\figfw{!ht}{width=1\linewidth}{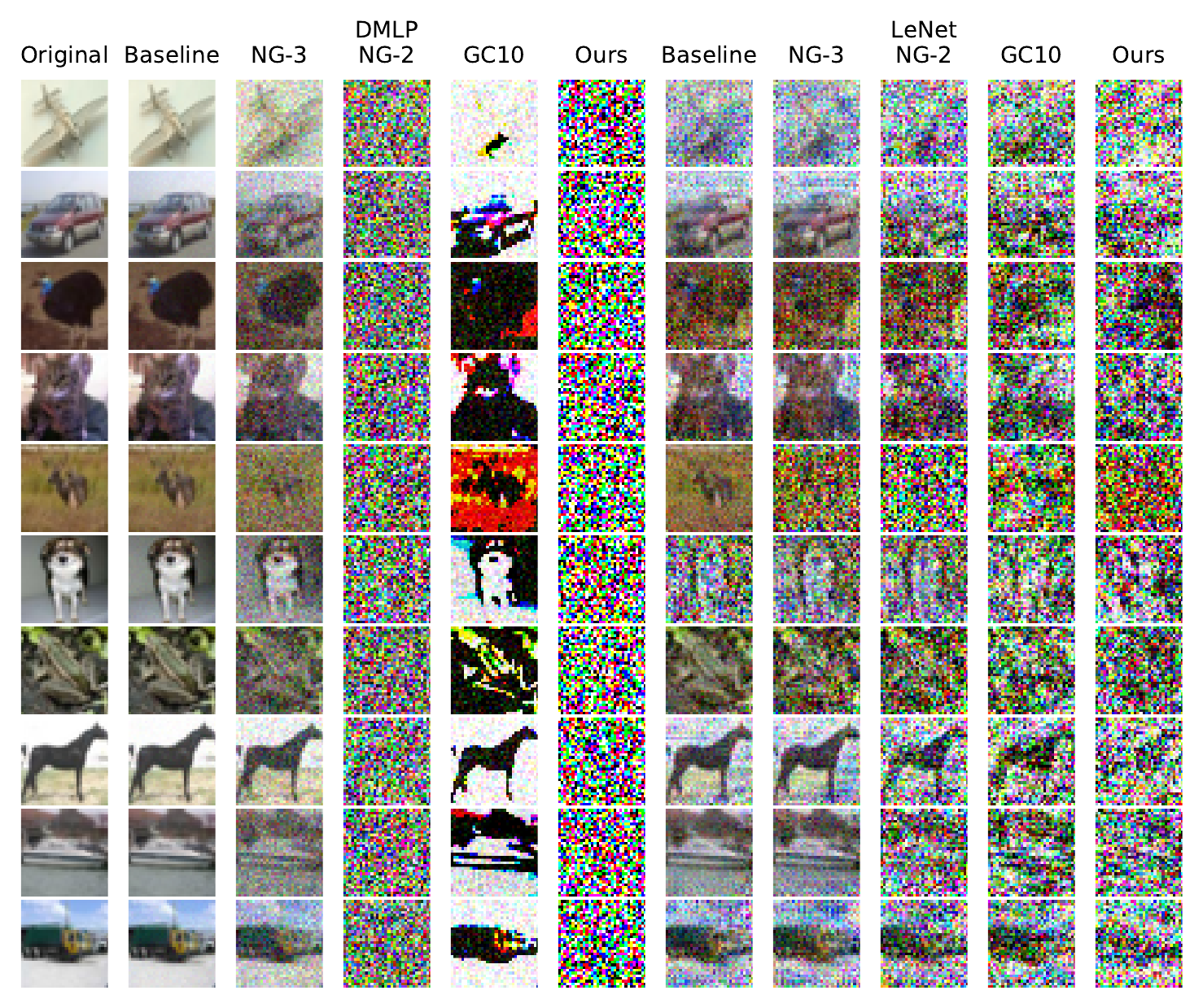}{fig:cif10rec}{Exemplary reconstruction results for all 10 classes of the CIFAR-10 dataset for the baseline DMLP and LeNet models and different defense mechanisms.}

\fig{!ht}{width=0.97\linewidth}{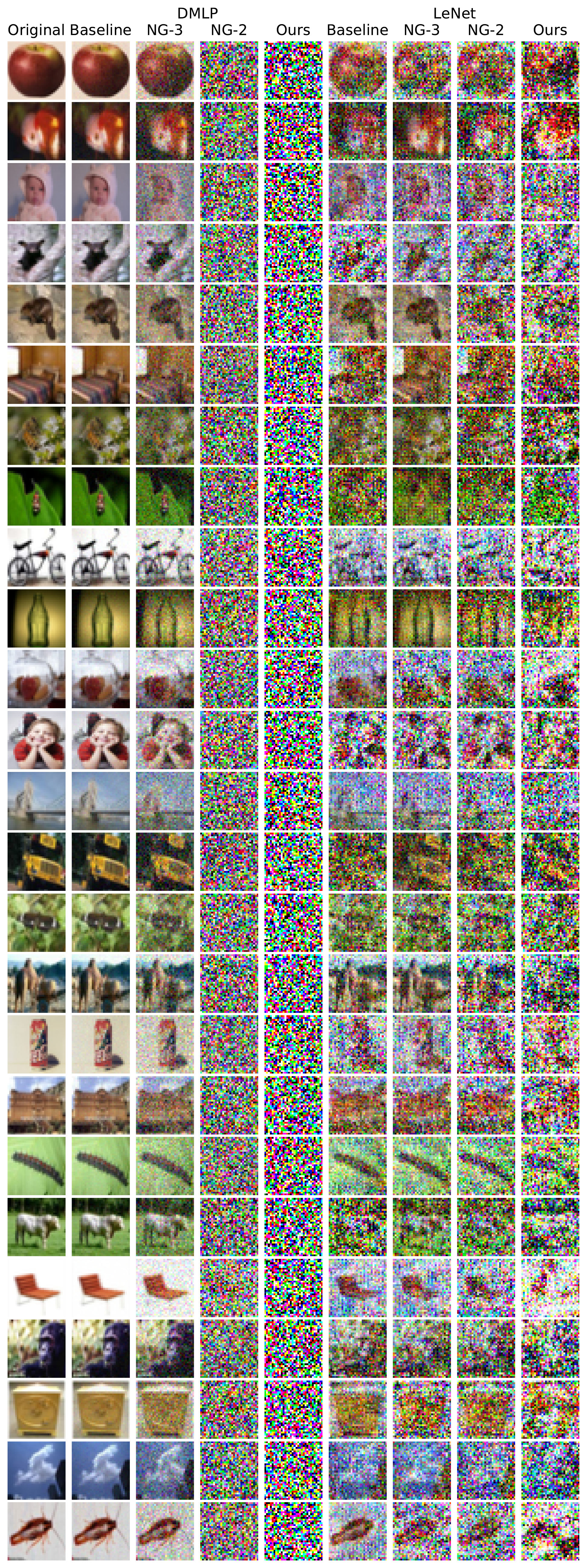}{fig:cif100_125}{Exemplary reconstruction results for classes 1-25 of the CIFAR-100 dataset for the baseline DMLP and LeNet models and different defense mechanisms.}
\fig{!ht}{width=0.97\linewidth}{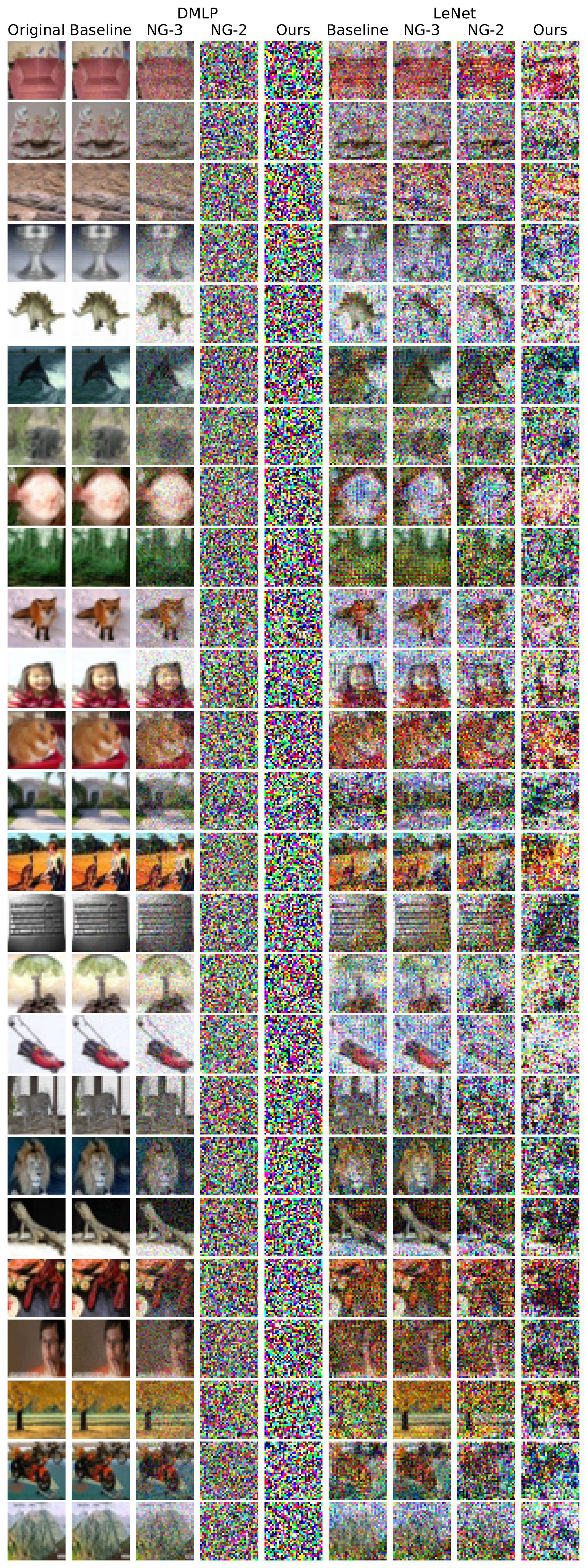}{fig:cif100_2650}{Exemplary reconstruction results for classes 26-50 of the CIFAR-100 dataset for the baseline DMLP and LeNet models and different defense mechanisms.}
\fig{!ht}{width=0.97\linewidth}{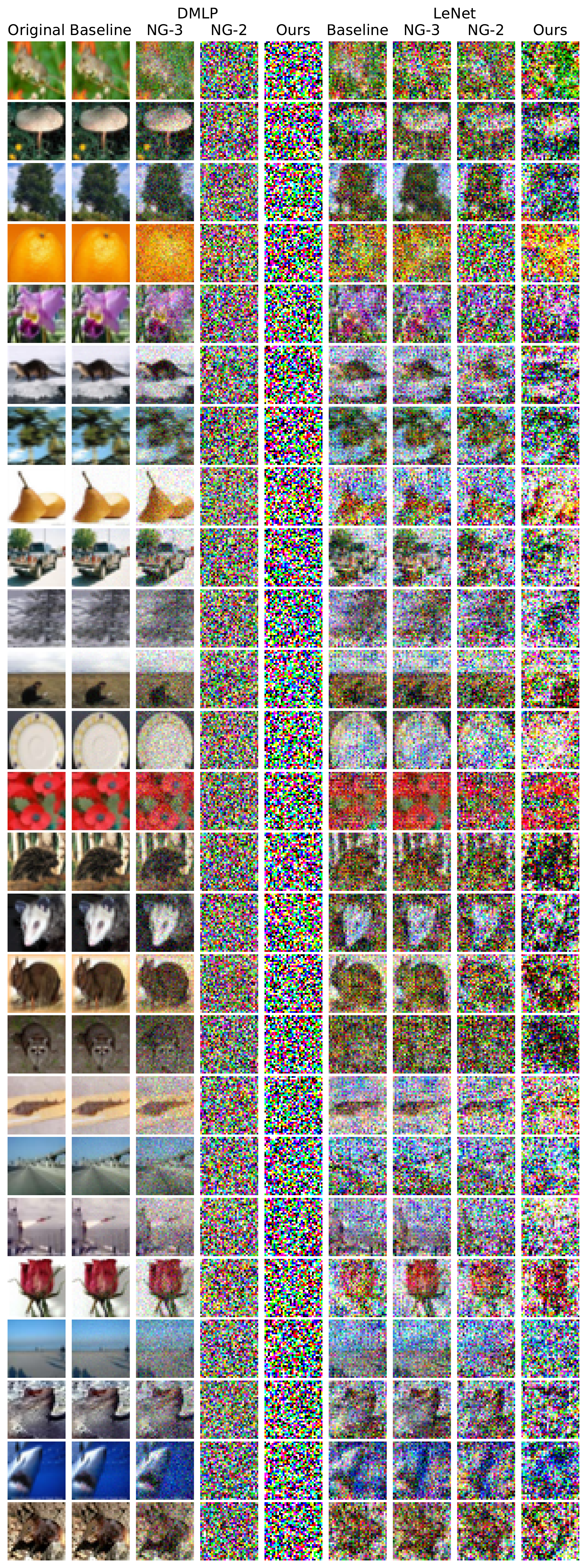}{fig:cif100_5175}{Exemplary reconstruction results for classes 51-75 of the CIFAR-100 dataset for the baseline DMLP and LeNet models and different defense mechanisms.}
\fig{!ht}{width=0.97\linewidth}{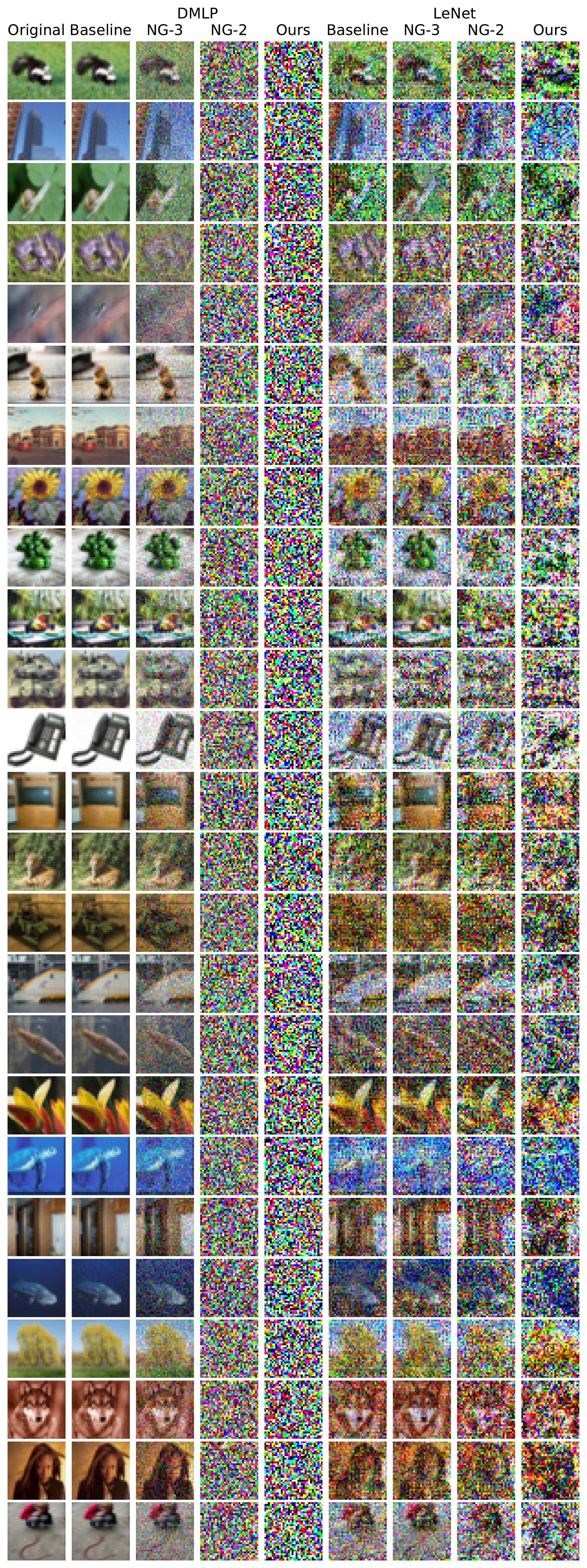}{fig:cif100_76100}{Exemplary reconstruction results for classes 76-100 of the CIFAR-100 dataset for the baseline DMLP and LeNet models and different defense mechanisms.}

\figfw{!ht}{width=.83\linewidth}{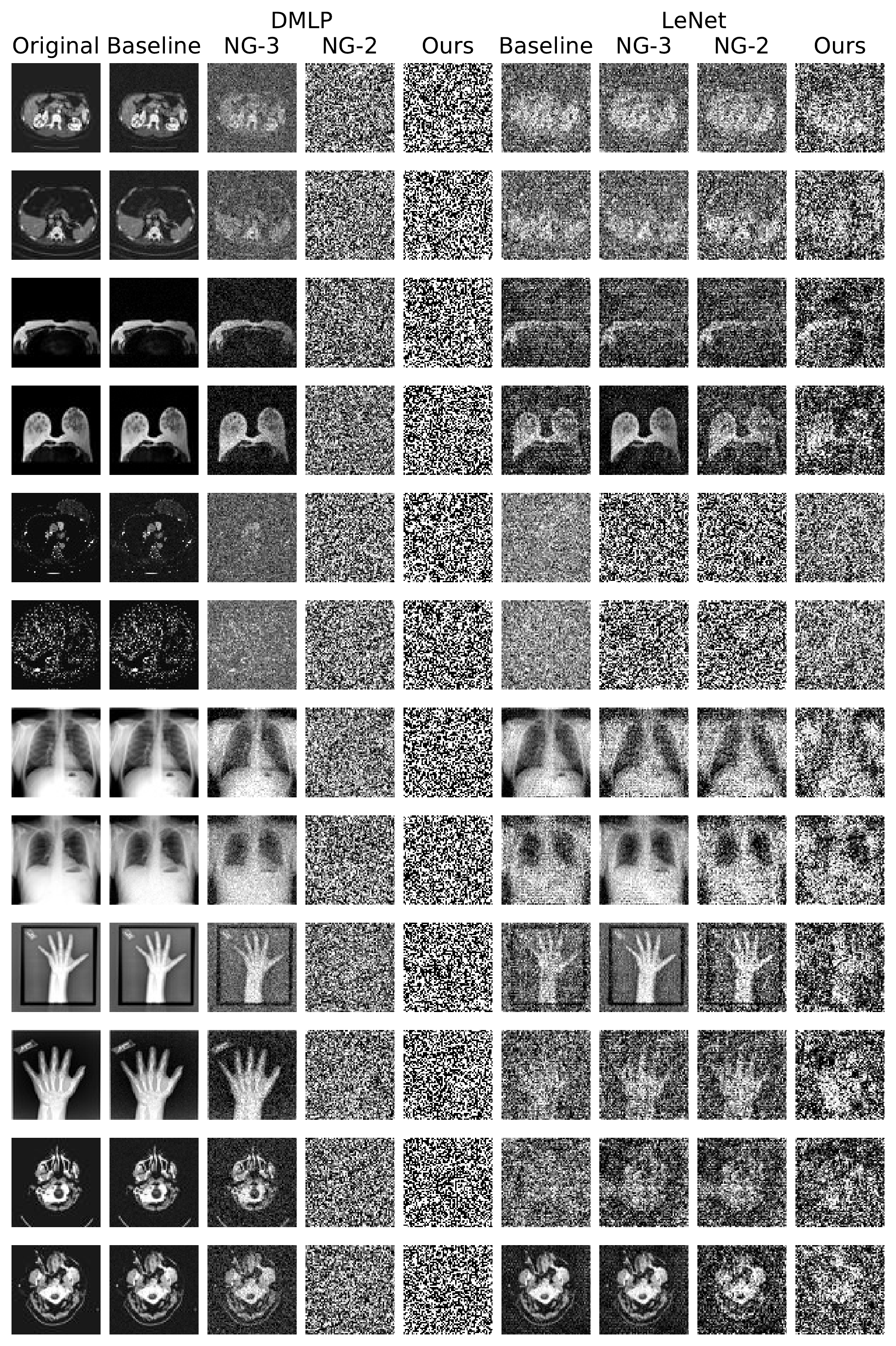}{fig:mmrec}{Exemplary reconstruction results for all 6 classes of the Medical MNIST dataset for the baseline DMLP and LeNet models and different defense mechanisms.}

\end{document}